\documentclass[runningheads]{llncs}
\usepackage{graphicx}
\usepackage{amsfonts}
\usepackage{amssymb}
\usepackage{siunitx}
\usepackage{float}
\usepackage{color}
\usepackage{array, makecell}
\usepackage{caption}
\usepackage{microtype}
\usepackage{float}
\usepackage{multirow,bigdelim,dcolumn,booktabs}
\usepackage{amsfonts}
\usepackage{makecell}
\usepackage{amsmath, bm}
\usepackage{color, colortbl}
\usepackage{subcaption}
\usepackage[ruled]{algorithm2e}
\SetKwInput{KwInput}{Input}
\SetKwInput{KwOutput}{Output}
\begin{document}
\title{Integrating Physiological Time Series and Clinical Notes with Transformer for Early Prediction of Sepsis}
\titlerunning{Early Sepsis Prediction}
%
\author{Yuqing Wang\inst{1\star} \and
Yun Zhao\inst{1\star} \and
Rachael Callcut\inst{2} \and
Linda Petzold\inst{1}}

\renewcommand{\thefootnote}{\fnsymbol{footnote}}
\footnotetext[1]{These two authors contributed  equally to this paper.}

\authorrunning{Y. Wang, Y. Zhao et al.}
%
\institute{Department of Computer Science, University of California, Santa Barbara \and
UC, Davis Health\\
\email{wang603@ucsb.edu, yunzhao@cs.ucsb.edu}\\
}
\maketitle              
\begin{abstract}
Sepsis is a leading cause of death in the Intensive Care Units (ICU). Early detection of sepsis is critical for patient survival. In this paper, we propose a multimodal Transformer model for early sepsis prediction, using the physiological time series data and clinical notes for each patient within $36$ hours of ICU admission. Specifically, we aim to predict sepsis using only the first $12$, $18$, $24$, $30$ and $36$ hours of laboratory measurements, vital signs, patient demographics, and clinical notes. We evaluate our model on two large critical care datasets: MIMIC-III and eICU-CRD.
The proposed method is compared with six baselines. In addition, ablation analysis and case studies are conducted to study the influence of each individual component of the model and the contribution of each data modality for early sepsis prediction. Experimental results demonstrate the effectiveness of our method, which outperforms competitive baselines on all metrics.

\keywords{Deep learning \and Transformer \and Sepsis.}
\end{abstract}

\section{Introduction}

Sepsis is a life-threatening organ dysfunction caused by a dysregulated host response to infection~\cite{singer2016third}, contributing to $30\%-50\%$ of inpatient mortality in the U.S~\cite{liu2014hospital}. The capability of early detection of sepsis allows for earlier interventions and treatment, thus improving patient outcomes. Following the widespread adoption of electronic health record (EHR) systems, researchers are particularly interested in using the EHR data to predict sepsis~\cite{desautels2016prediction,saqib2018early,masino2019machine}. 

One challenge of using EHR is that it stores both structured data (e.g., vital signs and laboratory measurements) and unstructured data (e.g., physician and nursing notes). Nevertheless, the heterogeneities across modalities increase the difficulty of performing sepsis prediction tasks. Thus, previous research works have focused on analyzing single data modality in isolation~\cite{saqib2018early,lipton2016directly,feng2020explainable}.  Structured physiological data can represent patients' true physiological signals.  However, in the case of sepsis, this data is incomplete and irregular due to urgency in the Intensive Care Units. Although unstructured medical notes can help understand patients' conditions more directly by capturing information regarding patients' symptom changes, it is insufficient to use notes alone to determine patients' status without support from physiological data.

To address the issues above, we propose a multimodal Transformer model that incorporates information from both physiological time series data and clinical notes for early prediction of sepsis. We use two large critical care datasets: the Medical Information Mart for Intensive Care III (MIMIC-III)~\cite{johnson2016mimic} and the eICU Collaborative Research Database (eICU-CRD)~\cite{pollard2018eicu}. Comprehensive experiments are conducted on the above two datasets to validate our approach, including performance comparison with baselines, ablation analysis, and case studies. Experimental results suggest that our proposed method outperforms six baselines on all metrics on both datasets. In addition, empirical ablation analysis and case studies indicate that each single modality contains unique information that is unavailable to the other modality. Hence, our model improves predictive performance by utilizing both physiological time series data and clinical notes.

The main contributions of this paper are highlighted as follows:

\begin{enumerate}
    \item[(1)] To the best of our knowledge, this is the first Transformer-based model that incorporates multivariate physiological time series data and clinical notes for early sepsis prediction.
    \item[(2)] Our experimental results indicate that both modalities complement each other. Thus, our method with both physiological data and clinical notes results in the best overall performance compared with unimodal methods. When using both modalities, our method outperforms competitive baselines on all metrics.
    \item[(3)] We perform attention mechanism visualization on clinical notes to improve the interpretability regarding the patients' status, which is not available in physiological data. In addition, distinctive distributions of physiological features between sepsis and non-sepsis patients demonstrate the unique information contained in physiological data but not in clinical notes.
\end{enumerate}

The remainder of this paper is organized as follows. Section~\ref{related_work} describes related work. The formal problem description is in Section~\ref{problem_def}. The proposed model is outlined in Section~\ref{methods}. Section~\ref{datasets} describes the datasets we use for evaluation. Experiments and results are discussed in Section~\ref{results}. Finally, our conclusions are presented in Section~\ref{conclusion}. 

\section{Related Work}\label{related_work}
In this section we review related work on clinical notes and physiological time series modeling, as well as multimodal methods in the clinical domain.

\subsection{Clinical Notes Modeling}
With the increasing availability of clinical notes over the past several years, there has been notable progress in understanding and using clinical text data to improve clinical prediction
outcomes. Natural language processing (NLP) and information retrieval techniques have been widely applied on different types of clinical tasks, such as  clinical relation extraction~\cite{patrick2010high},  de-identification of clinical notes~\cite{deleger2013large}, and clinical question answering~\cite{goodwin2016medical}. One common method for text representation is word embedding. In recent years, the appearance of the Transformer-based BERT~\cite{devlin2018bert} has offered an advantage over previous word embedding methods such as Word2Vec~\cite{mikolov2013distributed} and GloVe~\cite{pennington2014glove}  since it produces word representations that are dynamically informed by the words around them, which can effectively capture information from both the left and right contexts. In the clinical domain, BioBERT was pre-trained on PubMed abstracts and
articles and was able to better identify biomedical entities and boundaries than base BERT~\cite{lee2020biobert}. Base BERT and BioBERT have been further fine-tuned on the MIMIC-III dataset~\cite{johnson2016mimic} and released as ClinicalBERT and Clinical BioBERT, respectively \cite{alsentzer2019publicly}.

\subsection{Physiological Time Series Modeling}
Previous studies applied classical models such as Gaussian process (GP) and linear dynamical systems (LDS) to clinical time series modeling~\cite{liu2013modeling,liu2015clinical}. Given the growing availability of clinical data, recent studies demonstrate that RNN-based deep learning (DL) methods have become sought-after alternatives in clinical sequence modeling~\cite{saqib2018early,lipton2015learning}. More recently, an attention-based DL model has been proposed for clinical time series modeling~\cite{song2018attend}.

\subsection{Multimodal Methods in the Clinical Domain}
Multimodal representation learning is a fundamentally complex problem due to multiple sources of information~\cite{tsai2018learning}. Undeniably, multiple sources of data can provide complementary information, enabling more robust predictions~\cite{baltruvsaitis2018multimodal}. In the clinical domain, predictive models have been developed by integrating continuous monitoring data and discrete clinical event sequences~\cite{xu2018raim}. Combinations of multiple modalities such as clinical texts, procedures, medications, and laboratory measurements have shown improved performance on inpatient mortality, length of stay, and 30-day readmission prediction tasks~\cite{rajkomar2018scalable}. Unstructured clinical notes combined with structured measurements have been used for survival analysis of ICU trauma patients~\cite{zhao2021bertsurv}. 

\section{Problem Definition}\label{problem_def}
For a patient cohort consisting of $P$ patients, the multivariate physiological time series (MPTS) data associated with each patient can be expressed as 
$\{\bm{x}^{(1)},\bm{x}^{(2)},  \cdots, \bm{x}^{(M)}\} \in \mathbb{R}^{L \times M}$ with ${\bm{x}^{{(j)}^T}} = \{x^{(j)}_1, x^{(j)}_2, \cdots, x^{(j)}_L\}$. $M$ and $L$ represent the number of clinical features and the number of hours after admission, respectively. In addition, for each patient, sequences of clinical notes are used. The true label of each patient's clinical outcome is $y \in \{0,1\}$ (1 indicates sepsis and 0 indicates non-sepsis). Altogether, each of our datasets can be represented as $\{(\textbf{X}_i, \textbf{C}_i, {\rm Y}_i)|i = 1,2,\cdots,P\}$ where $\textbf{X}_i, \textbf{C}_i, {\rm Y}_i$ are the respective MPTS sequence, available clinical notes within $L$ hours, and the class label for patient $i$. We formulate sepsis prediction as a binary classification task, for which the goal is to learn a mapping:
\begin{align*}
    (\textbf{X}_i, \textbf{C}_i) \rightarrow {\rm Prob}( {{\rm Y}_i = 1|(\textbf{X}_i, \textbf{C}_i)}),
\end{align*}
where $i = 1,2,\cdots,P$.
In other words, MPTS data and clinical notes are used simultaneously to predict whether ICU patients admitted through the Emergency Department will develop sepsis. 


\section{Methods}\label{methods}
\begin{figure*}[ht]
\centering
\includegraphics[width=\linewidth]{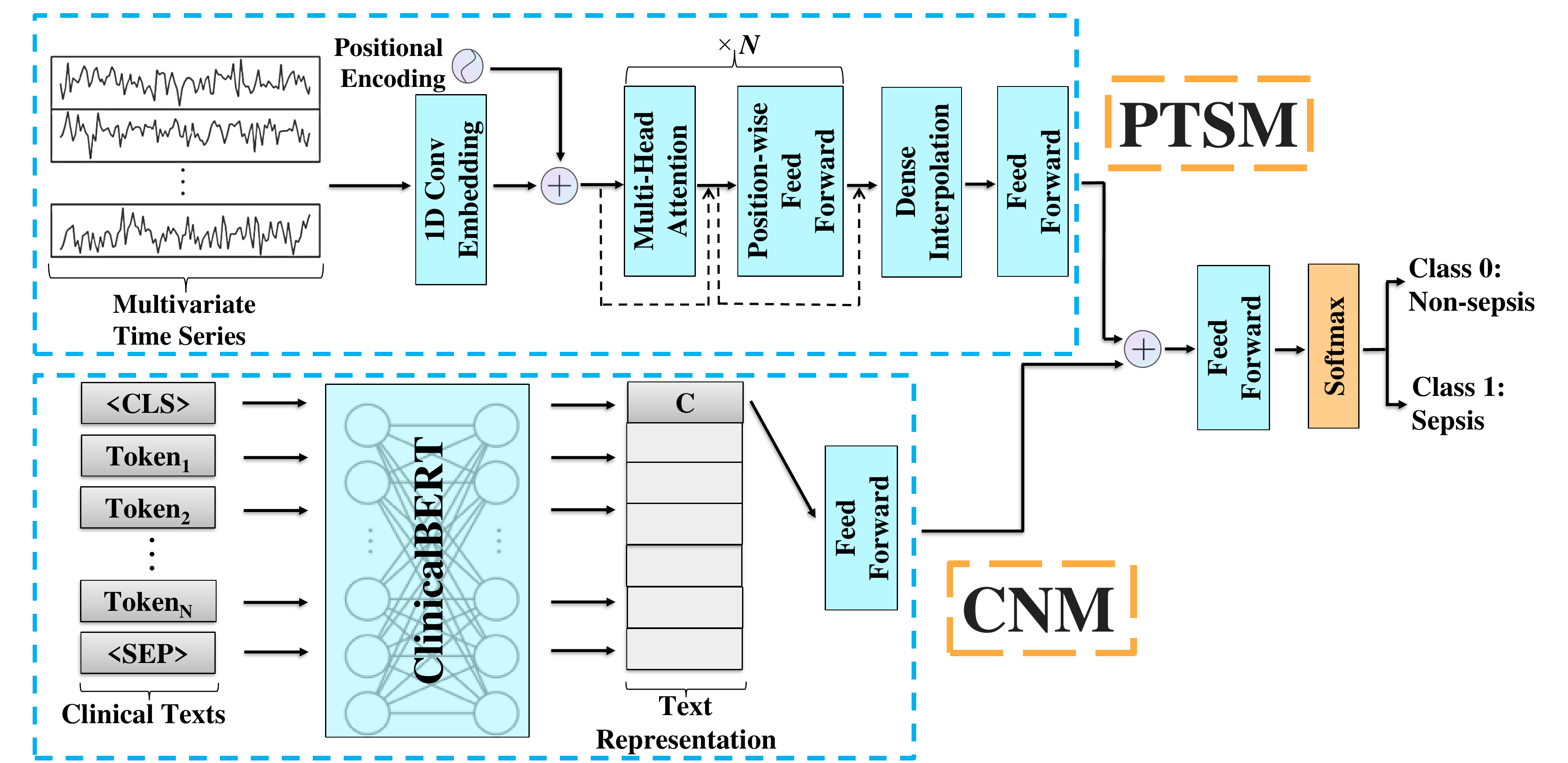}
\caption{An overview of the multimodal Transformer framework. The physiological time series model (PTSM) consists primarily of sequence embeddings, a stack of $N$ Transformer encoder layers (multi-head self-attention sublayer and position-wise FNN sublayer) with residual connection around each sublayer, dense interpolation, and FNN. The clinical notes model (CNM) consists of text representations using ClinicalBERT, and the output  $\langle$ CLS $\rangle$ representation is then used to feed into FNN. The output representations from PTSM and CNM are concatenated and fed into FNN, and the final Softmax layer is used for the binary classification task.}
\label{model}
\end{figure*}

In this section we propose the multimodal Transformer modeling framework. The model structure is illustrated in Figure~\ref{model}. 

\subsection{Clinical Notes Model (CNM)}
The CNM is composed of clinical text representations using ClinicalBERT~\cite{alsentzer2019publicly} and a feedforward neural network (FNN). The output $\langle$ CLS $\rangle$ representation following ClinicalBERT is fed into FNN.

\subsubsection{Transformer}
We begin by introducing Transformer's architecture~\cite{vaswani2017attention}, the foundation of Bidirectional Encoder Representations from Transformers (BERT), for which we use for clinical text representations. In Transformer~\cite{vaswani2017attention}, the self-attention mechanism enables the model to capture both short- and long-term dependencies, and different attention heads can learn different aspects of attention patterns. In the self-attention layer, an attention function maps a query $\textbf{Q}$ and a set of key-value pairs $\{\textbf{K,V}\}$ to an output $\textbf{O}$. Specifically, a multi-head self-attention sublayer simultaneously transforms the queries, keys and values into $H$ distinct and learnable linear projections, namely
\begin{align*}
    \textbf{Q}^h = \textbf{QW}_h^Q,\textbf{K}^h = \textbf{KW}_h^K,\textbf{V}^h = \textbf{VW}_h^V,
\end{align*}
where $\textbf{Q}^h, \textbf{K}^h, \textbf{V}^h$ are the respective query matrices, key matrices, and value matrices of the $h$-th attention head, with $h = \{1,2,\cdots,H\}$. Here, $\textbf{W}_h^Q,\textbf{W}_h^K \in \mathbb{R}^{d_{{\rm model}} \times d_k}$, $\textbf{W}_h^V \in \mathbb{R}^{d_{{\rm model}} \times d_v}$ denote learnable parameter matrices and $d_{{\rm model}}$ is the text 
embedding dimension. Next, $H$ attention functions are performed in parallel to produce a sequence of vector outputs:
\begin{align*}
    \textbf{O}^h &= \text{Attention}(\textbf{Q}^h, \textbf{K}^h, \textbf{V}^h) \\ &= \text{Softmax}\bigg(\frac{\textbf{Q}^h {\textbf{K}^h}^T}{\sqrt{d_k}} \bigg)\textbf{V}^h.
\end{align*}
Finally, the outputs $\textbf{O}^1, \textbf{O}^2,\cdots,\textbf{O}^H$ are concatenated and linearly projected again to produce the final representation.

\subsubsection{Text Representation with ClinicalBERT}
BERT is a pre-trained language representation based on the Transformer encoder architecture~\cite{devlin2018bert,vaswani2017attention}. BERT and its variants have exhibited outstanding performance in various NLP tasks. In medical contexts, ClinicalBERT develops clinically oriented word representations for clinical NLP tasks. Within ClinicalBERT, each token in clinical notes can be expressed as a sum of corresponding token embeddings, segment embeddings, and position embeddings. When feeding multiple sentences into ClinicalBERT, segment embeddings identify the sequence that a token is associated with and position embeddings of each token are a learned set of parameters corresponding to the token’s position in the input sequence~\cite{huang2019clinicalbert}. We use pre-trained ClinicalBERT for contextual representations of clinical notes.

\subsection{Physiological Time Series Model (PTSM)}
Inspired by the model architecture of Transformer~\cite{song2018attend,vaswani2017attention}, the PTSM is composed of sequence embeddings, positional encoding, a stack of $N$ identical Transformer encoder layers, and dense interpolation to incorporate temporal order.

\subsubsection{Input Embeddings}
In most NLP models, input embeddings are commonly used to map relatively low-dimensional vectors to high-dimensional vectors, which facilitate sequence modeling~\cite{kim-2014-convolutional}. For the same reason, a time sequence embedding is required to capture the dependencies among different features without considering the temporal information~\cite{song2018attend}. A 1D convolutional layer is employed to obtain the $K$-dimensional embeddings $(K > M)$ at each time step. 

\subsubsection{Positional Encoding}
In order to include the MPTS order information, we apply the same  sinusoidal functions for the positional encoding layer as ~\cite{vaswani2017attention} to encode the sequential information and add it to the input embeddings of the sequence.

\subsubsection{Transformer Encoder} 
We take advantage of the multi-head self-attention mechanism to capture dependencies of sequences. Similar to~\cite{vaswani2017attention}, we employ $8$ parallel attention heads. Following the attention output, a position-wise FNN is applied with two 1D convolutional layers with kernel size $1$, and a ReLU activation function in between. A residual connection is employed around each of the two sublayers.

\subsubsection{Dense Interpolation}
A concise representation of the output sequence from the Transformer encoder layer is needed since we do not make predictions at each time step~\cite{trask2015modeling}. A dense interpolation algorithm is applied on learned temporal representations for partial temporal order encoding. Given  MPTS data, the pseudocode to perform dense interpolation is shown in Algorithm $1$.
\begin{algorithm}[htbp]
\KwInput{time step $l$, time sequence length $L$, input embeddings $\textbf{e}_l$, interpolation coefficient $I$.}
\KwOutput{Dense representation $\textbf{z}$.}
 \For{l = $1$ \textbf{to} $L$}{
  e = $I$ $\times$ $l$ \slash $L$ \;
  \For{i = $1$ \textbf{to} I}{
   $r = pow(1-abs(e-i)/I,2)$ \;
   $\textbf{z}_i = \textbf{z}_i + r \times \textbf{e}_l$ \;
   }
 }
 \caption{Dense Interpolation}
\end{algorithm}
Let $\textbf{e}_l \in \mathbb{R}^{d_k}$ represent the intermediate representation following the Transformer encoder layers. The size of the interpolated embedding vector is $d_k \times I$, where $I$ is the interpolation coefficient. Algorithm $1$ mainly focuses on finding the contribution of $\textbf{e}_l$ to the position $i$ of the final representation $\textbf{z}$, denoted by $r$. At each time step $l$ , we obtain $e$, the relative position in the final vector representation $\textbf{z}$, and $r$ is computed as $r = (1 - \frac{|e-i|}{I})^2$. Finally, $\textbf{z}$ is obtained by matrix multiplication of $r$ and $\textbf{e}_l$ when we iterate through the time steps of a sequence.

\subsection{Incorporating PTSM and CNM}
The output representations from PTSM and CNM are concatenated, and the combined latent representation is fed into FNN. We use a Softmax layer as the final layer for the binary classification problem and the loss function is given by
\begin{align*}
    -(y \cdot {\rm log}(\hat{y})) + (1-y)\cdot {\rm log}(1 - \hat{y}),
\end{align*}
where $y$ and $\hat{y}$ are the true and predicted labels, respectively.

\section{Datasets}\label{datasets}
We use the MIMIC-III~\cite{johnson2016mimic} and eICU-CRD~\cite{pollard2018eicu} datasets to evaluate our method. MIMIC-III, a publicly available single-center clinical dataset, records $61,532$ ICU stays among $58, 976$ hospital admissions, including information on $46,520$ patients from Beth Israel Deaconess Medical Center between $2001$ and $2012$. The eICU-CRD, a multi-center dataset, consists of health data associated with over $200,000$ admissions to ICUs throughout continental United States between $2014$ and $2015$. Both datasets contain de-identified data, including patient demographics, vital signs, laboratory measurements, severity of illness, diagnosis, and clinical notes.

\subsection{Data Preprocessing Pipelines}
This section is divided into structured MPTS data preprocessing and unstructured clinical notes preprocessing, respectively.

\subsubsection{MPTS Data Preprocessing}
For both datasets, patient demographics, vital signs and laboratory measurements are extracted for ICU patients admitted through the Emergency Department. A list of clinically reasonable measurement ranges provided by~\cite{harutyunyan2019multitask} is used to remove outlier values. In total, we extracted $40$ and $38$ features from the MIMIC-III and eICU-CRD datasets, respectively. Since data was irregularly sampled, we resample the observation time into hourly bins for each feature. We use the mean value to determine feature values for which there are multiple records within an hour. Missing values are imputed by a combination of forward filling (i.e. using the value of the closest past bin 
with regard to the missing bin) and then backward filling (i.e. using the value of closest future bin 
with regard to the missing bin). In addition, we remove patients with hospital admission records of less than $12$ hours. We use only the first $T$ hours of MPTS data following patient admission for early sepsis prediction, where $T = 12,18,24,30,36$. For any patient whose measurement recording hours are less than $T$, his/her existing last-hour measurements are replicated to $T$. Otherwise, we truncate his/her measurement hours to $T$ such that the MPTS sequence length for all patients are guaranteed to be the same.

\subsubsection{Clinical Notes Preprocessing}
We use all the available clinical notes between hour $1$ and hour $36$ after ICU admission. If we use the first $T$ hours of MPTS data, then all the available notes up to $T$ hours are extracted for each patient. Over each interval of $T$ hours, for each patient we concatenate sequences of notes. Next, common text cleaning techniques are applied such as case normalization, stop words removal, and special characters removal are applied to clean the clinical notes. To
avoid potential label leakage, we remove sentences
containing “sepsis” or “septic”. Finally, the processed notes are fed into ClinicalBERT for text representations. 

\subsection{Sepsis Labeling}
We use the Angus criteria~\cite{angus2001epidemiology}, which is an International Classification of Diseases, Ninth Revision, Clinical Modification (ICD-9-CM) coding system, to identify sepsis for our datasets. Unlike other sepsis identification methods, it uses the final ICD diagnoses of organ failure and infection rather than feature values from the original datasets, to prevent data leakage issues~\cite{saqib2018early}. 

\subsection{Data Statistics}
After data preprocessing, we obtain a population of $18,625$ and $60,593$ ICU admissions for the MIMIC-III and eICU-CRD datasets, respectively. The sizes of positive and negative samples identified by the Angus criteria for each dataset are illustrated in Table $1$. Based on the ratio of negative and positive samples, our sepsis prediction task can be considered to be an imbalanced classification problem.
\vspace{-1cm}
\begin{table}[ht]
\centering
\caption{Sample sizes of two datasets.}
\begin{tabular}{c c c}
\Xhline{1.5pt}
Datasets  & \textbf{MIMIC-III} & \textbf{eICU-CRD}  \\ 
\hline
Total    & 18,625     & 60,593     \\ 
Negative & 11,655     & 55,926     \\ 
Positive & 6,970      & 4,667      \\
\Xhline{1.5pt}
\end{tabular}
\end{table}

\vspace{-1cm}
\section{Experiments and Results}\label{results}
Our experiments explore: $(1)$ the predictive performance of the multimodal Transformer model on the MIMIC-III and eICU-CRD datasets, $(2)$ the relative importance of individual components of the model through ablation analysis, and $(3)$ case studies on both clinical notes and MPTS data.

\subsection{Settings}
Both datasets are randomly split, with the training set and testing set of sizes $80\%$ and $20\%$, respectively. We
set aside $20\%$ of the training set for the validation set. Experiments are conducted using the first $12$, $18$, $24$, $30$ and $36$ hours of patient demographics, vital signs, laboratory measurements, and clinical notes for the Emergency Department patients on both datasets. All experiments were implemented in Pytorch~\cite{paszke2019pytorch} on one NVIDIA Tesla P100 GPU. We minimize the cross entropy loss with the Adam~\cite{kingma2014adam} optimizer for training. The hyperparameter search space for each dataset is listed in Table~\ref{hyperparameters}, where the hyperparameter values in bold indicate the optimal values found for our model using both modalities. Note that the batch size and the sequence length choice is limited by the available GPU memory. We perform grid search for hyperparameter optimization.
\begin{table}[htbp] 
\centering
\caption{Hyperparameter search space of our model on both datasets. Bold values are the
optimal values found using both modalities.}
\scalebox{0.8}{
\begin{tabular}{c c c} 
\Xhline{1.5pt}
Hyperparameters     & \textbf{MIMIC-III} & \textbf{eICU-CRD}  \\ 
\hline
learning rate       & [1e-4,\textbf{2e-5},3e-5,5e-5]      & [\textbf{1e-5},2e-5,3e-5,5e-5]     \\
dropout rate        & [\textbf{0.1},0.2,0.5]       & [\textbf{0.1},0.2,0.5]      \\
batch size          & [4,\textbf{8},12]   & [4,8,\textbf{12}]        \\
activation function & [\textbf{ReLU},SELU,GELU]      & [ReLU,SELU,\textbf{GELU}]       \\
training epochs     & [3,\textbf{4},5]         & [3,4,\textbf{5}]        \\
sequence length    & [256,\textbf{512}]       & [256,\textbf{512}]       \\
$\sharp$ of encoder layers $N$ & [3,4,5,\textbf{6}] & [3,4,5,\textbf{6}] \\
interpolation coefficient $I$ & [12,\textbf{24},32] & [12,24,\textbf{32}] \\
input embedding dim $K$ & [64,\textbf{128}] & [64,\textbf{128}] \\
class weight       & [0.5, \textbf{0.55}, 0.6, 0.65]     &  [0.0001,\textbf{0.0005},0.001]      \\
\Xhline{1.5pt}
\end{tabular}}
\label{hyperparameters}
\end{table}

\vspace{-1cm}
\subsection{Baselines and Evaluation Metrics}
We compare the performance of our model with the following six baselines where the first component (i.e., LSTM, BiLSTM, GRU) is commonly used for time series modeling and the second component (i.e., Word2Vec, FastText, ELMo) is commonly used for text representations in existing literature. Two components are integrated that support two modalities (i.e., time series and clinical notes).
\begin{itemize}
    \item \textbf{LSTM + CNM}~\cite{hochreiter1997long}
    \item \textbf{BiLSTM + CNM}~\cite{schuster1997bidirectional}
    \item \textbf{GRU + CNM}~\cite{cho2014properties}
    \item \textbf{PTSM + Word2Vec}~\cite{mikolov2013distributed}
    \item \textbf{PTSM + FastText}~\cite{bojanowski2017enriching}
    \item \textbf{PTSM + ELMo}~\cite{peters2018deep}
\end{itemize}
We evaluate our model performance in terms of area under the receiver operating characteristic curve (AUROC), F1 score, recall and precision, which are common metrics for imbalanced classification. In addition to the hyperparameters listed in Table~\ref{hyperparameters}, we fine-tune additional hyperparameters for LSTM (number of layers, hidden units), GRU (number of layers, hidden units), Word2Vec (window size, number of negative samples), FastText (maximum length of word n-gram, number of buckets), and ELMo (bidirectional and number of negative samples) as shown in Table~\ref{basehyperparameters}. The hyperparameter values in bold indicate the optimal values found for our baseline models using both modalities. The number of negative samples is based on the negative sampling algorithm. 

\begin{table}[ht] 
\centering
\caption{Hyperparameter search space of baselines on both datasets. Bold values are the optimal values found using both modalities.}
\scalebox{0.8}{
\begin{tabular}{c c c c} 
\Xhline{1.5pt}
& Hyperparameters     & \textbf{MIMIC-III} & \textbf{eICU-CRD}  \\ 
\hline
\multirow{3}{*}{LSTM}& $\sharp$ of layers & [1,2,\textbf{3},4] & [1,2,3,\textbf{4}]\\
    & hidden units & [100,150,\textbf{200}] & [100,150,\textbf{200}] \\
    & bidirectional & [\textbf{Yes}, No] & [\textbf{Yes}, No]\\
\hline
\multirow{2}{*}{GRU}& $\sharp$ of layers & [1,2,\textbf{3},4] & [1,2,\textbf{3},4]\\
    & hidden units & [100,150,\textbf{200}] & [100,150,\textbf{200}]\\
\hline
\multirow{3}{*}{Word2Vec} & window size & [5,10,\textbf{20}] & [5,\textbf{10},20]\\
    & \makecell{$\sharp$ of negative \\ samples} & [\textbf{10},15,20] & [10,\textbf{15},20] \\
\hline
\multirow{3}{*}{FastText}& \makecell{ max length of \\ word n-gram} & [2, \textbf{5}, 10] & [\textbf{2}, 5, 10]\\
    & $\sharp$ of buckets & [\textbf{1000},2000,3000] & [1000,\textbf{2000},3000]\\
\hline  
\multirow{3}{*}{ELMo}& bidirectional & [\textbf{Yes}, No] & [\textbf{Yes}, No]\\
    & \makecell{$\sharp$ of negative \\ samples} & [10,
    15, \textbf{20}, 30] & [10, 15, 20, \textbf{30}]\\

\Xhline{1.5pt}
\end{tabular}}
\label{basehyperparameters}
\end{table}

\vspace{-1cm}
\subsection{Results}
The results of our method and all baselines using both modalities on two datasets are shown in Table~\ref{mimic_result} and \ref{eicu_result}, respectively. We can see that our method outperforms all baselines on both datasets on all metrics regardless of hours we use. Compared to LSTM and GRU, PTSM benefited from its self-attention mechanism. Specifically, PTSM has direct access to all of the available data 
in parallel, which leaves no room for information loss. Furthermore, compared to Word2Vec and FastText, CNM (ClinicalBERT) provides dynamic contextualized word representations instead of static embeddings, which brings about flexible text representations. For ELMo, since it is based on BiLSTM, it may not be able to deal with long-term dependencies as well as Transformer-based CNM. In general, all models performed better when supplied with available MPTS data and clinical notes covering more hours. 

\subsection{Ablation Analysis}
To further study the influence of each individual component of our proposed method, we conduct ablation experiments to investigate the influence of individual model components with different data inputs. The results of ablation analysis on both datasets are presented in Table~\ref{mimic_abl} and \ref{eicu_abl}, respectively. First, we consider the performance of applying MPTS data on PTSM only and clinical notes on CNM only. As can be seen from Table~\ref{abl}, in general, the model with input of solely MPTS data has better performance than that of solely clinical notes. Next, we utilize both data modalities with only hour $1$ MPTS data (admission measurements) and available clinical notes within $T$ hours since admission where $T = 12,18,24,30,36$. When using both modalities, they can bring about comparable results with those of using MPTS data only. Finally, in terms of our full model using full MPTS data and clinical notes, the performance improves with the available data covering more hours by a margin of $4.3\%-8.5\%$ on AUROC, $4.1\%-7.3\%$ on F1 score, $3.3\%-9.1\%$ on precision, and $3.0\%-7.9\%$ on recall compared with the ``best” model performance when using partial data. The ablation analysis suggests that both MPTS data and clinical notes complement and benefit each other and thus the model with both modalities 
has better performance than the model with single modality.

\vspace{-0.5cm}
\begin{table}[htbp] 
\caption{Performance comparison for the MIMIC-III and eICU-CRD datasets between the proposed method and six baselines. Hours represent all the data available including MPTS and clinical notes after admission. Experiments are conducted $5$ times with different random seeds. The results are shown in the format of mean and standard deviation.}
\begin{subtable}[h]{0.49\textwidth}
\centering
\setlength{\tabcolsep}{1mm}
\resizebox{\textwidth}{!}{
\begin{tabular}{c| c c c c c c} 
\Xhline{1.5pt}
& \textbf{Hours} & $12$ & $18$ & $24$ & $30$ & $36$ \\
\hline
\multirow{4}{*}{\makecell{Baseline 1: \\LSTM + CNM}} & AUROC & \ 0.854 $\pm$ 0.009 \ & \ 0.867 $\pm$ 0.008 \ & \ 0.875 $\pm$ 0.008 \  & \ 0.878 $\pm$ 0.009  \ & \ 0.884 $\pm$ 0.008 \ \\
    & F1 Score  &  \ 0.846 $\pm$ 0.006 \  & \ 0.852 $\pm$ 0.007 \  &  \ 0.856 $\pm$ 0.007 \  &  \ 0.857 $\pm$ 0.006 \ & \ 0.861 $\pm$ 0.006 \ \\
    & Precision  &  \ 0.797 $\pm$ 0.006 \ & \ 0.799 $\pm$ 0.007 \    & \ 0.801 $\pm$ 0.007 \   & \ 0.802 $\pm$ 0.006 \ & \ 0.807 $\pm$ 0.006 \ \\
    & Recall & \ 0.901 $\pm$ 0.006 \   & \ 0.913 $\pm$ 0.007 \   &  \ 0.918 $\pm$ 0.007 \ &  \ 0.921 $\pm$ 0.006 \ & \ 0.923 $\pm$ 0.006 \ \\
\hline
\multirow{4}{*}{\makecell{Baseline 2: \\BiLSTM + CNM}} & AUROC & \ 0.861 $\pm$ 0.008 \ & \ 0.869 $\pm$ 0.008 \  & \ 0.878 $\pm$ 0.009 \ & \ 0.886 $\pm$ 0.009 \ & \ 0.890 $\pm$ 0.008 \ \\
    & F1 Score  & \ 0.853 $\pm$ 0.009 \  &  \ 0.858 $\pm$ 0.007  \ & \ 0.862 $\pm$ 0.008 \ & \ 0.865 $\pm$ 0.009 \ & \ 0.869 $\pm$ 0.008\ \\
    & Precision  & \ 0.803 $\pm$ 0.009 \   & \ 0.808 $\pm$ 0.007 \ & \ 0.811 $\pm$ 0.008 \ & \ 0.813 $\pm$ 0.009 \ & \ 0.816 $\pm$ 0.008 \ \\
    & Recall & \ 0.909 $\pm$ 0.009 \ &  \ 0.914 $\pm$ 0.007  \ & \ 0.920 $\pm$ 0.008 \ & \ 0.924 $\pm$ 0.009 \ & \ 0.930 $\pm$ 0.008 \ \\
\hline
\multirow{4}{*}{\makecell{Baseline 3: \\GRU + CNM}} & AUROC & \ 0.849 $\pm$ 0.011 \ & \ 0.856 $\pm$ 0.012 \ & \ 0.864 $\pm$ 0.011 \ & \ 0.871 $\pm$ 0.012 \ & \ 0.876 $\pm$ 0.010 \ \\
    & F1 Score  &  \ 0.842 $\pm$ 0.009 \ & \ 0.844 $\pm$ 0.010 \ & \ 0.848 $\pm$ 0.009 \ & \ 0.851 $\pm$ 0.011 \ & \ 0.853 $\pm$ 0.012 \ \\
    & Precision  & \ 0.795 $\pm$ 0.009 \ & \ 0.797 $\pm$ 0.010 \ & \ 0.802 $\pm$ 0.009 \ & \ 0.805 $\pm$ 0.011 \ & \ 0.806 $\pm$ 0.012 \  \\
    & Recall & \ 0.896 $\pm$ 0.009 \ & \ 0.898 $\pm$ 0.010 \ & \ 0.900 $\pm$ 0.009 \ & \ 0.903 $\pm$ 0.011 \ & \ 0.906 $\pm$ 0.012 \ \\
\hline
\multirow{4}{*}{\makecell{Baseline 4: \\ PTSM + Word2Vec}}& AUROC & \ 0.838 $\pm$ 0.008 \  & \ 0.851 $\pm$ 0.007 \  & \ 0.859 $\pm$ 0.007 \ & \ 0.863 $\pm$ 0.007 \   & \ 0.872 $\pm$ 0.008 \  \\
& F1 Score & \ 0.830 $\pm$ 0.009 \ & \ 0.836 $\pm$ 0.008 \    & \ 0.848 $\pm$ 0.007 \    & \ 0.851 $\pm$ 0.008 \    & \ 0.855 $\pm$ 0.009 \  \\
& Precision & \ 0.792 $\pm$ 0.009 \   & \ 0.794 $\pm$ 0.008 \    & \ 0.798 $\pm$ 0.007 \     & \ 0.800 $\pm$ 0.008 \    & \ 0.801 $\pm$ 0.009 \ \\
& Recall & \ 0.872 $\pm$ 0.009 \   & \ 0.882 $\pm$ 0.008 \    & \ 0.905 $\pm$ 0.007 \    & \ 0.910 $\pm$ 0.008 \   &  \ 0.916 $\pm$ 0.009 \  \\
\hline
\multirow{4}{*}{\makecell{Baseline 5: \\PTSM + FastText}}& AUROC &  \ 0.859 $\pm$ 0.007 \ & \ 0.868 $\pm$ 0.009 \ & \ 0.875 $\pm$ 0.012 \ & \ 0.883 $\pm$ 0.010 \ & \ 0.889 $\pm$ 0.009 \ \\
    & F1 Score &  \ 0.851 $\pm$ 0.008 \ & \ 0.854 $\pm$ 0.009 \ & \ 0.859 $\pm$ 0.007 \ & \ 0.861 $\pm$ 0.007 \ & \ 0.865 $\pm$ 0.009 \ \\
    & Precision & \ 0.801 $\pm$ 0.008 \ & \ 0.804 $\pm$ 0.009 \ & \ 0.809 $\pm$ 0.007 \ & \ 0.811 $\pm$ 0.007 \ & \ 0.815 $\pm$ 0.009 \ \\
    & Recall & \ 0.907 $\pm$ 0.008 \ & \ 0.911 $\pm$ 0.009 \ & \ 0.915 $\pm$ 0.007 \ & \ 0.918 $\pm$ 0.007 \ & \ 0.922 $\pm$ 0.009 \ \\
\hline
\multirow{4}{*}{\makecell{Baseline 6: \\PTSM + ELMo}}& AUROC &  \ 0.863 $\pm$ 0.005 \ & \ 0.871 $\pm$ 0.007 \ & \ 0.880 $\pm$ 0.006 \ & \ 0.889 $\pm$ 0.007 \ & \ 0.892 $\pm$ 0.006 \ \\
    & F1 Score &  \ 0.854 $\pm$ 0.006 \ & \ 0.859 $\pm$ 0.007 \ & \ 0.863 $\pm$ 0.006 \ & \ 0.867 $\pm$ 0.008 \ & \ 0.870 $\pm$ 0.007 \\\
    & Precision & \ 0.805 $\pm$ 0.006 \ & \ 0.810 $\pm$ 0.007 \ & \ 0.814 $\pm$ 0.006 \ & \ 0.817 $\pm$ 0.008 \ & \ 0.819 $\pm$ 0.007 \ \\
    & Recall & \ 0.910 $\pm$ 0.006 \ & \ 0.915 $\pm$ 0.007 \ & \ 0.918 $\pm$ 0.006 \ & \ 0.923 $\pm$ 0.008 \ & \ 0.928 $\pm$ 0.007 \  \\
\hline
\multirow{4}{*}{\makecell{Ours: \\PTSM + CNM}} & AUROC  & \ \textbf{0.902 $\pm$ 0.004} \   & \ \textbf{0.910 $\pm$ 0.005} \  & \ \textbf{0.917 $\pm$ 0.005} \   & \ \textbf{0.923 $\pm$ 0.004} \   & \ \textbf{0.928 $\pm$ 0.004} \ \\
    & F1 Score &  \ \textbf{0.881 $\pm$ 0.005} \  & \ \textbf{0.887 $\pm$ 0.006} \   & \ \textbf{0.894 $\pm$ 0.004} \  &  \ \textbf{0.907 $\pm$ 0.005} \  &  \ \textbf{0.910 $\pm$ 0.004} \ \\
    & Precision & \ \textbf{0.839 $\pm$ 0.005} \    & \ \textbf{0.845 $\pm$ 0.006} \   & \ \textbf{0.852 $\pm$ 0.004} \   &  \ \textbf{0.866 $\pm$ 0.005} \  & \ \textbf{0.869 $\pm$ 0.004} \  \\
    & Recall & \ \textbf{0.928 $\pm$ 0.005} \   &  \ \textbf{0.933 $\pm$ 0.006} \  &  \ \textbf{0.940 $\pm$ 0.004} \  & \ \textbf{0.951 $\pm$ 0.005} \    & \ \textbf{0.955 $\pm$ 0.004} \ \\   
\Xhline{1.5pt}
\end{tabular}}
\caption{Comparison results on the MIMIC-III testing set.}
\label{mimic_result}
\end{subtable}
\hfill
\begin{subtable}[h]{0.49\textwidth}
\centering
\setlength{\tabcolsep}{1mm}
\resizebox{\textwidth}{!}{
\begin{tabular}{c| c c c c c c c} 
\Xhline{1.5pt}
& \textbf{Hours} & $12$ & $18$ & $24$ & $30$ & $36$ \\
\hline
\multirow{4}{*}{\makecell{Baseline 1: \\LSTM + CNM}} & AUROC & \ 0.796 $\pm$ 0.012 \ & \ 0.801 $\pm$ 0.010 \ & \ 0.816 $\pm$ 0.009 \ & \ 0.827 $\pm$ 0.010 \ & \ 0.830 $\pm$ 0.011 \ \\
    & F1 Score  & \ 0.787 $\pm$ 0.009 \ & \ 0.792 $\pm$ 0.008 \ & \ 0.794 $\pm$ 0.008 \ & \ 0.796 $\pm$ 0.009 \ & \ 0.798 $\pm$ 0.010 \  \\
    & Precision  & \ 0.773 $\pm$ 0.009 \ & \ 0.779 $\pm$ 0.008 \ & \ 0.782 $\pm$ 0.008 \ & \ 0.783 $\pm$ 0.009 \ & \ 0.786 $\pm$ 0.010\\
    & Recall & 0.802 $\pm$ 0.009 & 0.805 $\pm$ 0.008 & 0.806 $\pm$ 0.008 & 0.809 $\pm$ 0.009 & 0.810 $\pm$ 0.010  \\
\hline
\multirow{4}{*}{\makecell{Baseline 2: \\BiLSTM + CNM}} & AUROC & \ 0.802 $\pm$ 0.012 \ & \ 0.809 $\pm$ 0.011 \ & \ 0.825 $\pm$ 0.009 \ & \ 0.833 $\pm$ 0.009 \  & \ 0.851 $\pm$ 0.009  \ \\
    & F1 Score  & \ 0.790 $\pm$ 0.008 \  & \ 0.801 $\pm$ 0.009 \    & \ 0.813 $\pm$ 0.009 \ &  \ 0.820 $\pm$ 0.008\ & \ 0.827 $\pm$ 0.008 \ \\
    & Precision  & \ 0.778 $\pm$ 0.008 \ & \ 0.781 $\pm$ 0.009 \ & \ 0.785 $\pm$ 0.009 \    & \ 0.787 $\pm$ 0.008 \   & \ 0.794 $\pm$ 0.008 \ \\
    & Recall & \ 0.802 $\pm$ 0.008 \  & \ 0.821 $\pm$ 0.009 \  & \ 0.844 $\pm$ 0.009 \   &  \ 0.855 $\pm$ 0.008 \ & \ 0.863 $\pm$ 0.008 \ \\
\hline
\multirow{4}{*}{\makecell{Baseline 3: \\GRU + CNM}} & AUROC & \ 0.791 $\pm$ 0.007 \ & \ 0.800 $\pm$ 0.008 \  & \ 0.813 $\pm$ 0.008  \ & \ 0.824 $\pm$ 0.009 \ & \ 0.829 $\pm$ 0.008 \ \\
    & F1 Score  & \ 0.773 $\pm$ 0.008 \   & \ 0.782 $\pm$ 0.008\ & \ 0.786 $\pm$ 0.007 \ & \ 0.793 $\pm$ 0.006\ & \ 0.797 $\pm$ 0.007\ \\
    & Precision  & \ 0.776 $\pm$ 0.008 \   & \ 0.780 $\pm$ 0.008 \ & \ 0.783 $\pm$ 0.007\ & \ 0.787 $\pm$ 0.006\ & \ 0.792 $\pm$ 0.007\ \\
    & Recall & \ 0.770 $\pm$ 0.008 \ &  \ 0.784 $\pm$ 0.008 \ & \ 0.789 $\pm$ 0.007 \ & \ 0.799 $\pm$ 0.006 \ & \ 0.803 $\pm$ 0.007 \ \\
\hline
\multirow{4}{*}{\makecell{Baseline 4: \\PTSM + Word2Vec}}& AUROC & \ \ 0.787 $\pm$ 0.009 \  & \ 0.796 $\pm$ 0.009 \  & \ 0.811 $\pm$ 0.008 \ & \ 0.824 $\pm$ 0.010 \   & \ 0.834 $\pm$ 0.009\  \\
    & F1 Score & \ 0.784 $\pm$ 0.008 \ & \ 0.788 $\pm$ 0.007 \    & \ 0.804 $\pm$ 0.007\    & \ 0.813 $\pm$ 0.008 \    & \ 0.821 $\pm$ 0.008 \  \\
    & Precision & \ 0.778 $\pm$ 0.008 \   & \ 0.780 $\pm$ 0.007 \    & \ 0.784 $\pm$ 0.007 \     & \ 0.787 $\pm$ 0.008 \    & \ 0.792 $\pm$ 0.008 \  \\
    & Recall & \ 0.791 $\pm$ 0.008 \   & \ 0.796 $\pm$ 0.007 \    & \ 0.824 $\pm$ 0.007 \    & \ 0.841 $\pm$ 0.008 \   &  \ 0.852 $\pm$ 0.008 \  \\
\hline
\multirow{4}{*}{\makecell{Baseline 5: \\PTSM + FastText}}& AUROC & \ 0.814 $\pm$ 0.012 \ & \ 0.826 $\pm$ 0.011\ & \ 0.838 $\pm$ 0.011 \ & \ 0.846 $\pm$ 0.010 \ & \ 0.852 $\pm$ 0.011 \   \\
    & F1 Score & \ 0.807 $\pm$ 0.009 \    & \ 0.815 $\pm$ 0.010 \    & \ 0.822 $\pm$ 0.010 \    & \ 0.828 $\pm$ 0.009 \    & \ 0.834 $\pm$ 0.010\ \\
    & Precision & \ 0.802 $\pm$ 0.009 \    & \ 0.805 $\pm$ 0.010 \     & \ 0.808 $\pm$ 0.010 \    & \ 0.810 $\pm$ 0.009 \    & \  0.813 $\pm$ 0.010 \ \\
    & Recall & \ 0.813 $\pm$ 0.009 \  & \ 0.826 $\pm$ 0.010 \    & \ 0.837 $\pm$ 0.010 \   &  \ 0.848 $\pm$ 0.009 \  & \ 0.856 $\pm$ 0.010 \  \\
\hline
\multirow{4}{*}{\makecell{Baseline 6: \\PTSM + ELMo}}& AUROC & \ 0.812 $\pm$ 0.008 \  & \ 0.821 $\pm$ 0.009 \ & \ 0.832 $\pm$ 0.007\   & \ 0.844 $\pm$ 0.008 \ & \ 0.849 $\pm$ 0.009\  \\
    & F1 Score & \ 0.808 $\pm$ 0.007 \    & \ 0.815 $\pm$ 0.006 \    & \ 0.819 $\pm$ 0.007 \    & \ 0.824 $\pm$ 0.008  \    & \ 0.830 $\pm$ 0.007 \ \\
     & Precision & \ 0.803 $\pm$ 0.007 \    & \ 0.807 $\pm$ 0.006 \    & \ 0.809 $\pm$ 0.007 \   &  \ 0.811 $\pm$ 0.008 \  & \ 0.812 $\pm$ 0.007 \  \\
    & Recall & \ 0.814 $\pm$ 0.007  \    & \ 0.823 $\pm$ 0.006\     & \ 0.829 $\pm$ 0.007  \    & \ 0.837 $\pm$ 0.008 \    & \ 0.849 $\pm$ 0.007 \ \\
\hline
\multirow{4}{*}{\makecell{Ours: \\PTSM + CNM}} & AUROC  & \ \ \textbf{0.845 $\pm$ 0.006} \   &  \ \textbf{0.852 $\pm$ 0.005} \  & \ \textbf{0.861 $\pm$ 0.005} \   & \ \textbf{0.873 $\pm$ 0.006} \   &  \ \textbf{0.882 $\pm$ 0.004} \ \\
    & F1 Score &  \ \textbf{0.833 $\pm$ 0.005} \   &  \ \textbf{0.840 $\pm$ 0.004} \  & \ \textbf{0.845 $\pm$ 0.004} \   &  \ \textbf{0.851 $\pm$ 0.004} \  & \ \textbf{0.857 $\pm$ 0.003} \  \\
    & Precision & \ \textbf{0.802 $\pm$ 0.005} \   & \ \textbf{0.807 $\pm$ 0.004} \   & \ \textbf{0.809 $\pm$ 0.004} \   & \ \textbf{0.814 $\pm$ 0.004} \    &  \ \textbf{0.818 $\pm$ 0.003} \  \\
    & Recall & \ \textbf{0.866 $\pm$ 0.005} \  & \ \textbf{0.875 $\pm$ 0.004} \   &  \ \textbf{0.884 $\pm$ 0.004} \  & \ \textbf{0.892 $\pm$ 0.004} \  &  \ \textbf{0.900 $\pm$ 0.003} \ \\ 
\Xhline{1.5pt}
\end{tabular}}
\caption{Comparison results on the eICU-CRD testing set.}
\label{eicu_result}
\end{subtable}
\end{table}

\vspace{-1cm}
\begin{table}[htbp]
\caption{Ablation analysis on the influence of different components in our model for the MIMIC-III and eICU-CRD datasets. Experiments are conducted $5$ times with different random seeds. The results are shown in the format of mean and standard deviation. Note that hour $1$ MPTS indicates that only initial measurements are considered as input instead of a series of measurements. Also, the case that hour $1$ clinical notes (i.e. admission notes) with increasing available MPTS data is not considered since the available notes for each patient at the initial time is limited.}
\begin{subtable}[h]{0.49\textwidth}
\centering
\setlength{\tabcolsep}{1mm}
\resizebox{\textwidth}{!}{
\begin{tabular}{c| c c c c c c} 
\Xhline{1.5pt}
& \textbf{Hours} & $12$ & $18$ & $24$ & $30$ & $36$  \\ 
\hline
\multirow{4}{*}{\makecell{MPTS on\\PTSM only}}
&\ AUROC \ & \ 0.827 $\pm$ 0.009 \   & \ 0.835 $\pm$ 0.008 \  & \ 0.839 $\pm$ 0.010 \   & \ 0.842 $\pm$ 0.007 \   &  \ 0.846 $\pm$ 0.008 \    \\
&\ F1 Score \  & \ 0.822 $\pm$ 0.006 \   &  \ 0.830 $\pm$ 0.007 \  & \ 0.831 $\pm$ 0.006 \   & \ 0.837 $\pm$ 0.008 \   &  \ 0.838 $\pm$ 0.007 \    \\
&\ Precision \  & \ 0.777 $\pm$ 0.006 \   & \ 0.784 $\pm$ 0.007 \   & \ 0.771 $\pm$ 0.007 \    &  \ 0.793 $\pm$ 0.008 \  & \ 0.778 $\pm$ 0.007\     \\
&\ Recall  \ &  \ 0.872 $\pm$ 0.006 \  & \ 0.882 $\pm$ 0.007 \   & \ 0.900 $\pm$ 0.006 \   & \ 0.887 $\pm$ 0.008 \   & \ 0.907 $\pm$ 0.007 \     \\ 
\hline
\multirow{4}{*}{\makecell{Clinical Notes \\ on CNM only}}
&\ AUROC  \ & \ 0.790 $\pm$ 0.008 \   &  \ 0.797 $\pm$ 0.007\  & \ 0.806 $\pm$ 0.008 \    &  \ 0.812 $\pm$ 0.009\  &  \ 0.831 $\pm$ 0.007 \    \\
&\ F1 Score \ & \ 0.776 $\pm$ 0.007 \   & \ 0.789 $\pm$ 0.007\    & \ 0.799 $\pm$ 0.009 \   &  \ 0.804 $\pm$ 0.008 \  & \ 0.823 $\pm$ 0.007 \     \\
&\ Precision \ & \ 0.749 $\pm$ 0.008 \   & \ 0.740 $\pm$ 0.007 \   & \ 0.784 $\pm$ 0.009\   & \ 0.782 $\pm$ 0.009 \   & \ 0.792 $\pm$ 0.007 \     \\
&\ Recall  \ & \ 0.806 $\pm$ 0.007 \   &  \ 0.846 $\pm$ 0.007\  & \ 0.814 $\pm$ 0.009 \   & \ 0.828 $\pm$ 0.008 \   &  \ 0.856 $\pm$ 0.007 \    \\ 
\hline
\multirow{4}{*}{\makecell{Hour $1$ MPTS \\ on PTSM \\ + CNM}}
& \ AUROC    \ & \ 0.836 $\pm$ 0.011 \  & \ 0.839 $\pm$ 0.010 \ & \ 0.846 $\pm$ 0.008 \  & \ 0.862 $\pm$ 0.009 \   &  \ 0.871 $\pm$ 0.009 \    \\
& \ F1 Score   \ &  \ 0.830 $\pm$ 0.009 \  & \ 0.831 $\pm$ 0.008 \   &  \ 0.833 $\pm$ 0.008 \  &  \ 0.847 $\pm$ 0.009 \  &  \ 0.863 $\pm$ 0.010 \    \\
& \ Precision  \ &  \ 0.795 $\pm$ 0.009 \  & \ 0.787 $\pm$ 0.008 \   & \ 0.789 $\pm$ 0.008\   &  \ 0.794 $\pm$ 0.009 \  & \ 0.808 $\pm$ 0.010 \     \\
& \ Recall  \  & \ 0.869 $\pm$ 0.009 \   & \ 0.880 $\pm$ 0.008 \   & \ 0.883 $\pm$ 0.008 \   &  \ 0.907 $\pm$ 0.009 \  & \ 0.927 $\pm$ 0.010 \     \\ 
\hline
\multirow{4}{*}{\makecell{Ours: \\PTSM + CNM}} & AUROC  & \ \textbf{0.902 $\pm$ 0.004} \   & \ \textbf{0.910 $\pm$ 0.005} \  & \ \textbf{0.917 $\pm$ 0.005} \   & \ \textbf{0.923 $\pm$ 0.004} \   & \ \textbf{0.928 $\pm$ 0.004} \ \\
    & F1 Score &  \ \textbf{0.881 $\pm$ 0.005} \  & \ \textbf{0.887 $\pm$ 0.006} \   & \ \textbf{0.894 $\pm$ 0.004} \  &  \ \textbf{0.907 $\pm$ 0.005} \  &  \ \textbf{0.910 $\pm$ 0.004} \ \\
    & Precision & \ \textbf{0.839 $\pm$ 0.005} \    & \ \textbf{0.845 $\pm$ 0.006} \   & \ \textbf{0.852 $\pm$ 0.004} \   &  \ \textbf{0.866 $\pm$ 0.005} \  & \ \textbf{0.869 $\pm$ 0.004} \  \\
    & Recall & \ \textbf{0.928 $\pm$ 0.005} \   &  \ \textbf{0.933 $\pm$ 0.006} \  &  \ \textbf{0.940 $\pm$ 0.004} \  & \ \textbf{0.951 $\pm$ 0.005} \    & \ \textbf{0.955 $\pm$ 0.004} \ \\ 
\hline
\end{tabular}}
       \caption{Ablation analysis results on the MIMIC-III testing set.}
       \label{mimic_abl}
    \end{subtable}
    \hfill
    \begin{subtable}[h]{0.49\textwidth}
\centering
\setlength{\tabcolsep}{1mm}
\resizebox{\textwidth}{!}{
\begin{tabular}{c| c c c c c c } 
\Xhline{1.5pt}
& \textbf{Hours} & $12$ & $18$ & $24$ & $30$ & $36$  \\ 
\hline
\multirow{4}{*}{\makecell{MPTS on\\PTSM only}}
&\ AUROC  \ & \ 0.782 $\pm$ 0.006 \   & \ 0.788 $\pm$ 0.007 \   &  \ 0.793 $\pm$ 0.008 \  &  \ 0.796 $\pm$ 0.009 \  &  \ 0.817 $\pm$ 0.008 \    \\
&\ F1 Score \ & \ 0.773 $\pm$ 0.005 \   &  \ 0.776 $\pm$ 0.006 \  & \ 0.780 $\pm$ 0.006 \   & \ 0.781 $\pm$ 0.007 \   & \ 0.799 $\pm$ 0.006 \     \\
&\ Precision  \ & \ 0.766 $\pm$ 0.005 \   &  \ 0.750 $\pm$ 0.006 \  & \ 0.733 $\pm$ 0.007 \    &  \ 0.727 $\pm$ 0.007\  &  \ 0.757 $\pm$ 0.006 \    \\
&\ Recall \  &  \ 0.780 $\pm$ 0.005 \  & \ 0.803 $\pm$ 0.006\   &  \ 0.833 $\pm$ 0.006 \  &  \ 0.844 $\pm$ 0.007 \  & \ 0.847 $\pm$ 0.006 \     \\ 
\hline
\multirow{4}{*}{\makecell{Clinical Notes \\ on CNM only}}
&\ AUROC  \  & \ 0.724 $\pm$ 0.008 \ & \ 0.733 $\pm$ 0.010 \ & \ 0.748 $\pm$ 0.009 \   &  \ 0.756 $\pm$ 0.007 \  & \ 0.778 $\pm$ 0.008 \  \\
&\ F1 Score  \ & \ 0.717 $\pm$ 0.007\ & \ 0.721 $\pm$ 0.006 \ & \ 0.736 $\pm$ 0.006  \ & \ 0.742 $\pm$ 0.007  \ & \ 0.761 $\pm$ 0.008\ \\
&\ Precision  \ & \ 0.695 $\pm$ 0.007\ & \ 0.692 $\pm$ 0.007 \  &  \ 0.704 $\pm$ 0.006\   & \ 0.708 $\pm$ 0.007 \    & \ 0.719 $\pm$ 0.008\  \\
&\ Recall  \ & \ 0.740 $\pm$ 0.007 \ & \ 0.752 $\pm$ 0.006 \  &  \ 0.771 $\pm$ 0.006 \  & \ 0.780 $\pm$ 0.007 \   & \ 0.808 $\pm$ 0.007 \   \\ 
\hline
\multirow{4}{*}{\makecell{Hour $1$ MPTS \\ on PTSM \\ + CNM}}
&\ AUROC \ & \ 0.794 $\pm$ 0.011 \   &  \ 0.801 $\pm$ 0.009\  &   \ 0.814 $\pm$ 0.008\ & \ 0.831 $\pm$ 0.009 \   &  \ 0.846 $\pm$ 0.008 \    \\
&\ F1 Score \  &  \ 0.787 $\pm$ 0.008 \  & \ 0.792 $\pm$ 0.007 \   & \ 0.805 $\pm$ 0.007 \   & \ 0.817 $\pm$ 0.008 \    & \ 0.823 $\pm$ 0.008 \     \\
&\ Precision \   & \ 0.765 $\pm$ 0.008 \   & \ 0.768 $\pm$ 0.007 \   & \ 0.777 $\pm$ 0.007\   & \ 0.786 $\pm$ 0.008 \   & \ 0.792 $\pm$ 0.008 \     \\
& \ Recall \ & \ 0.811 $\pm$ 0.008 \   & \ 0.818 $\pm$ 0.007\   &  \ 0.836 $\pm$ 0.007 \  & \ 0.851 $\pm$ 0.008\    & \ 0.857 $\pm$ 0.008 \     \\ 
\hline
\multirow{4}{*}{\makecell{Ours: \\PTSM + CNM}} & AUROC  & \ \ \textbf{0.845 $\pm$ 0.006} \   &  \ \textbf{0.852 $\pm$ 0.005} \  & \ \textbf{0.861 $\pm$ 0.005} \   & \ \textbf{0.873 $\pm$ 0.006} \   &  \ \textbf{0.882 $\pm$ 0.004} \ \\
    & F1 Score &  \ \textbf{0.833 $\pm$ 0.005} \   &  \ \textbf{0.840 $\pm$ 0.004} \  & \ \textbf{0.845 $\pm$ 0.004} \   &  \ \textbf{0.851 $\pm$ 0.004} \  & \ \textbf{0.857 $\pm$ 0.003} \  \\
    & Precision & \ \textbf{0.802 $\pm$ 0.005} \   & \ \textbf{0.807 $\pm$ 0.004} \   & \ \textbf{0.809 $\pm$ 0.004} \   & \ \textbf{0.814 $\pm$ 0.004} \    &  \ \textbf{0.818 $\pm$ 0.003} \  \\
    & Recall & \ \textbf{0.866 $\pm$ 0.005} \  & \ \textbf{0.875 $\pm$ 0.004} \   &  \ \textbf{0.884 $\pm$ 0.004} \  & \ \textbf{0.892 $\pm$ 0.004} \  &  \ \textbf{0.900 $\pm$ 0.003} \ \\ 
\hline
\end{tabular}}
       \caption{Ablation analysis results on the eICU-CRD testing set.}
       \label{eicu_abl}
    \end{subtable}
     \label{abl}
\end{table}

\begin{figure}[htbp]
    \centering
    \subfloat[Respiratory Rates ]{\includegraphics[width=0.235\textwidth,]{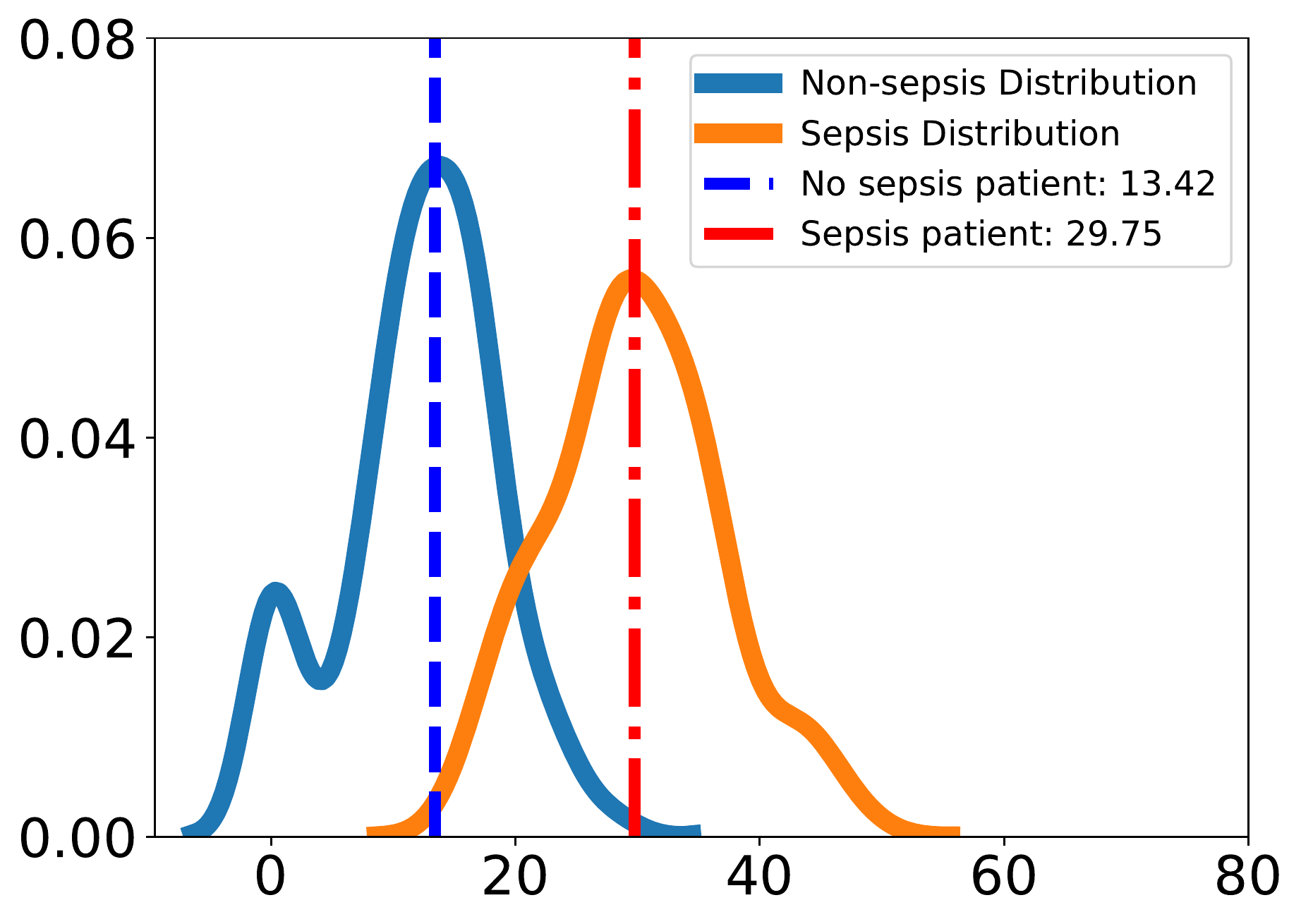}}
    \hfill
    \subfloat[Diastolic Blood Pressure]{\includegraphics[width=0.235\textwidth, ]{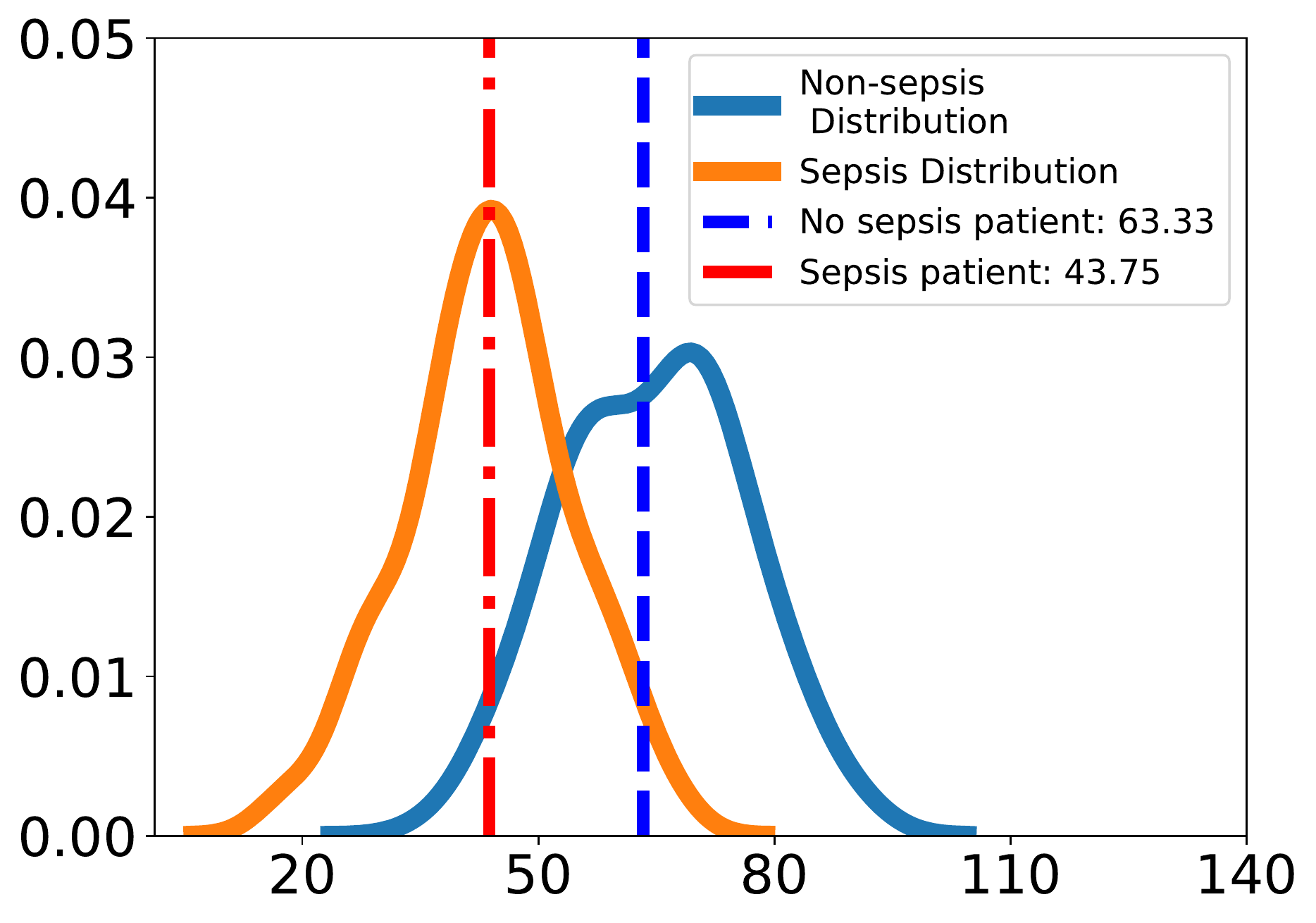}}
    \hfill
    \subfloat[Bicarbonate Levels]{\includegraphics[width=0.235\textwidth, ]{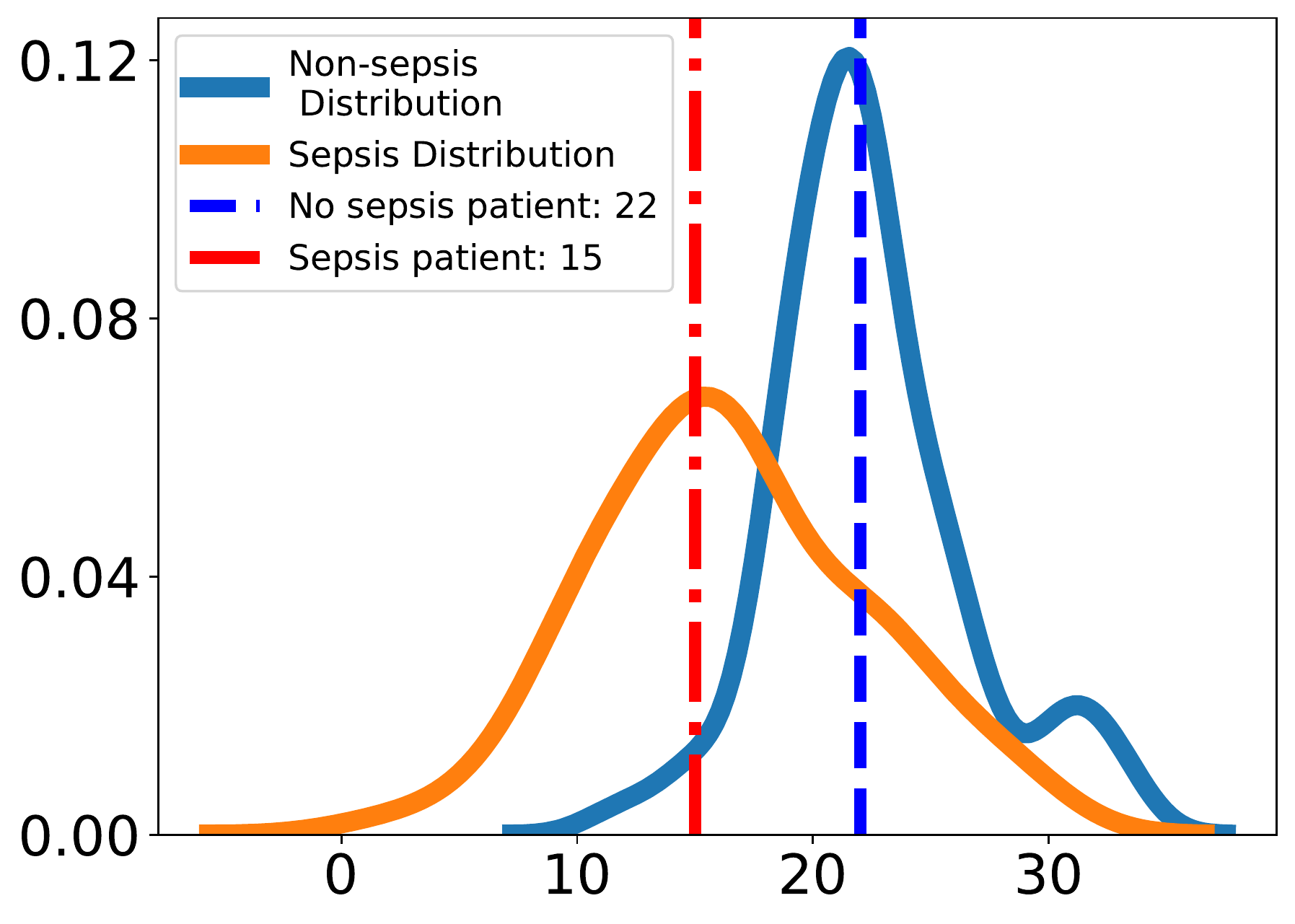}}
    \hfill
    \subfloat[Oxygen Saturation]{\includegraphics[width=0.235\textwidth, ]{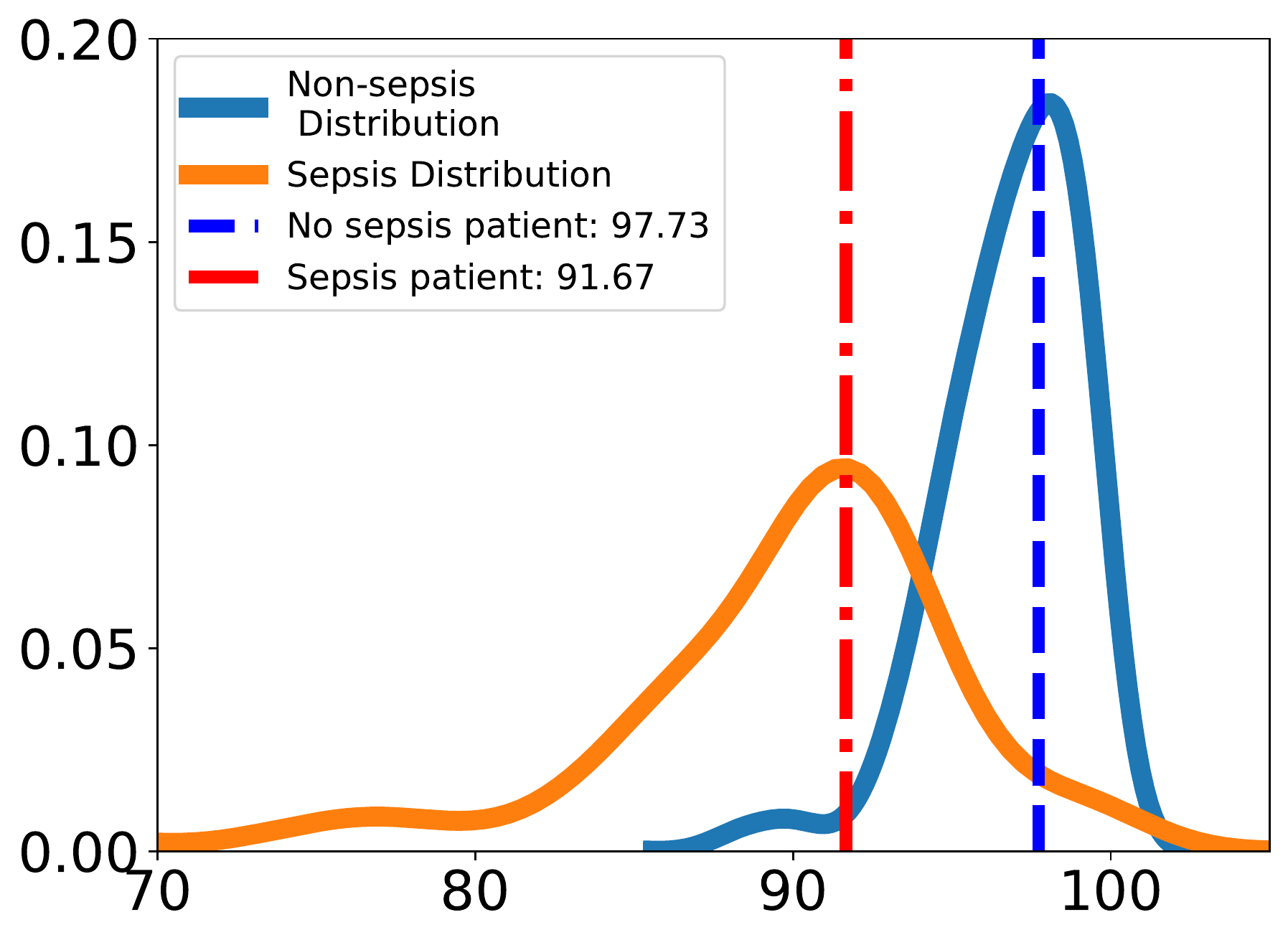}}
  \caption{Density plots of features. The blue and orange curves are density curves of corresponding features. The blue curve represents no sepsis, and the orange represents sepsis. The dashed vertical lines shows the two patients' feature values.}
  \label{fig_case}
\end{figure}

\begin{figure}[htbp]
    \centering
    \subfloat[patient with sepsis ]{\includegraphics[width=0.235\textwidth]{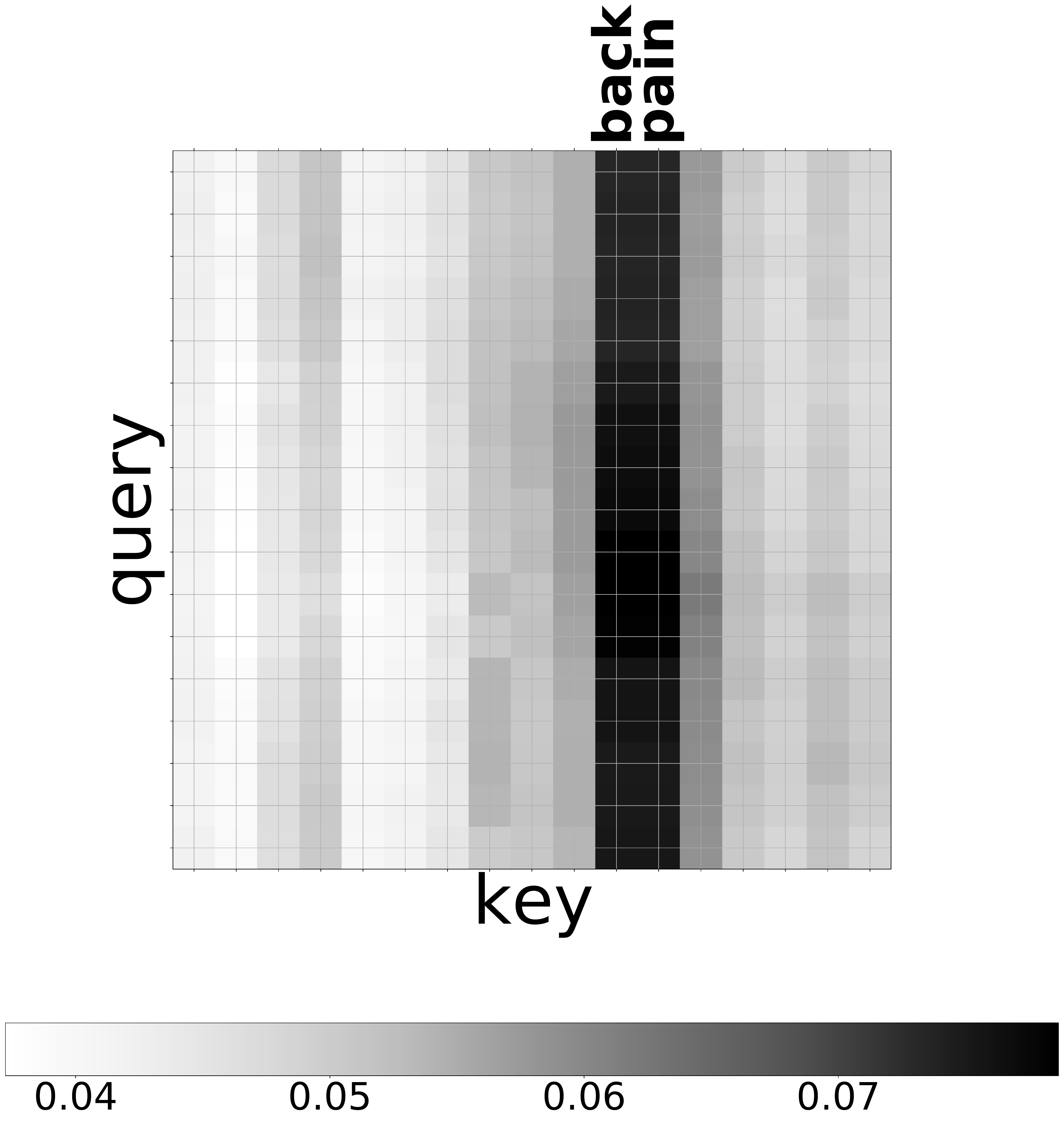}}
    \hfill
    \subfloat[patient with sepsis]{\includegraphics[width=0.235\textwidth]{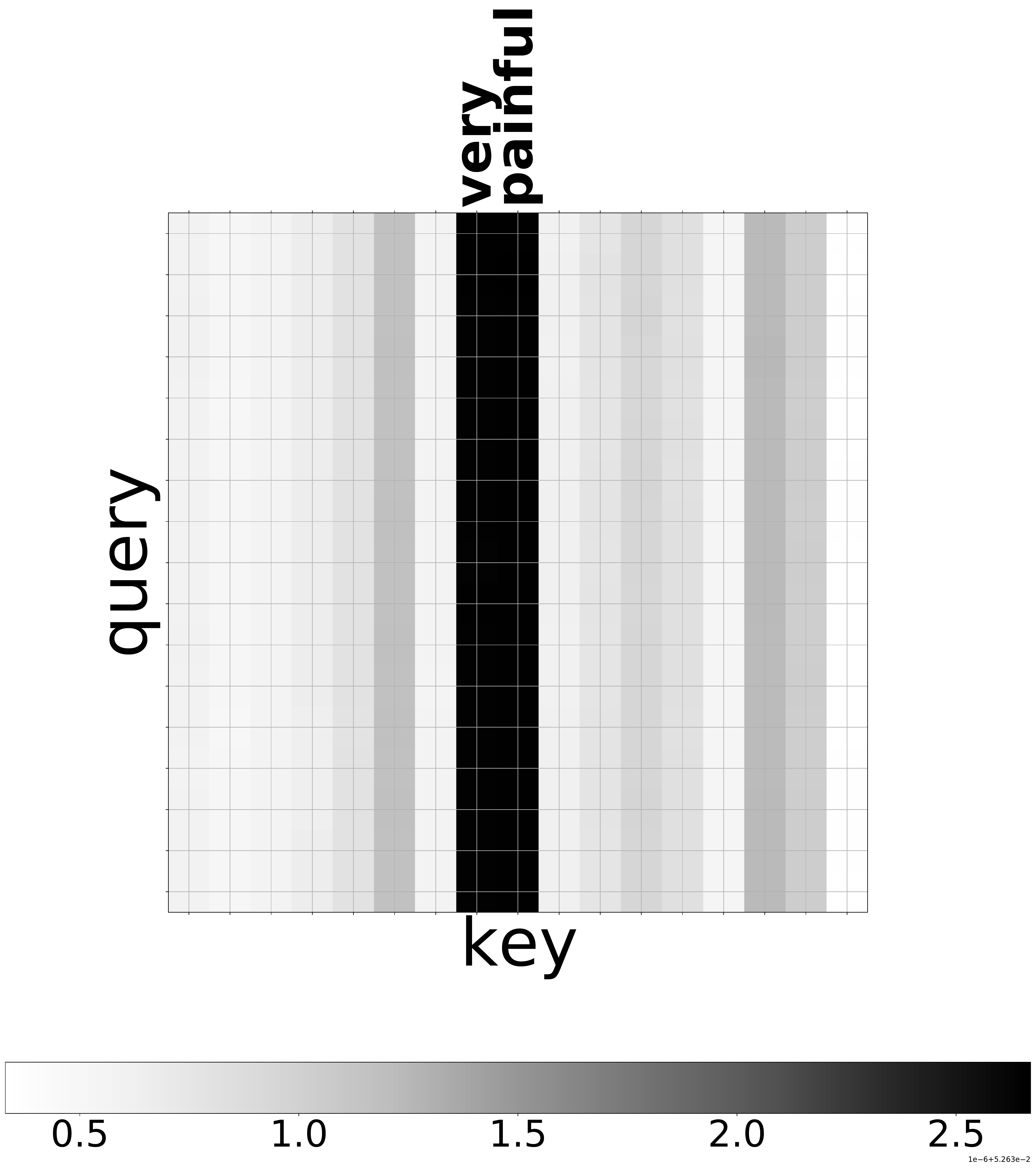}}
    \quad
    \subfloat[patient without sepsis]{\includegraphics[width=0.235\textwidth]{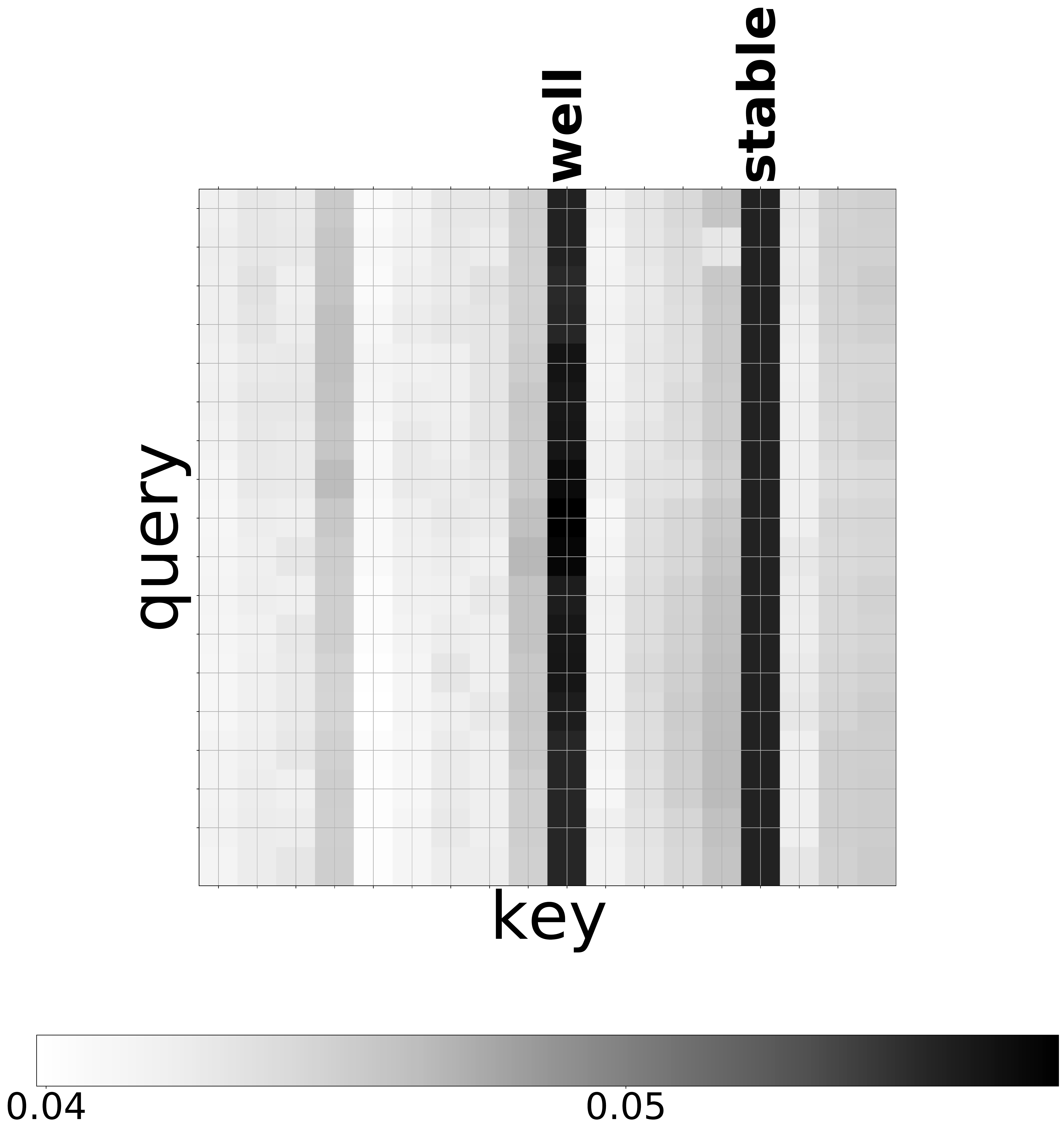}}
    \hfill
    \subfloat[patient without sepsis]{\includegraphics[width=0.235\textwidth]{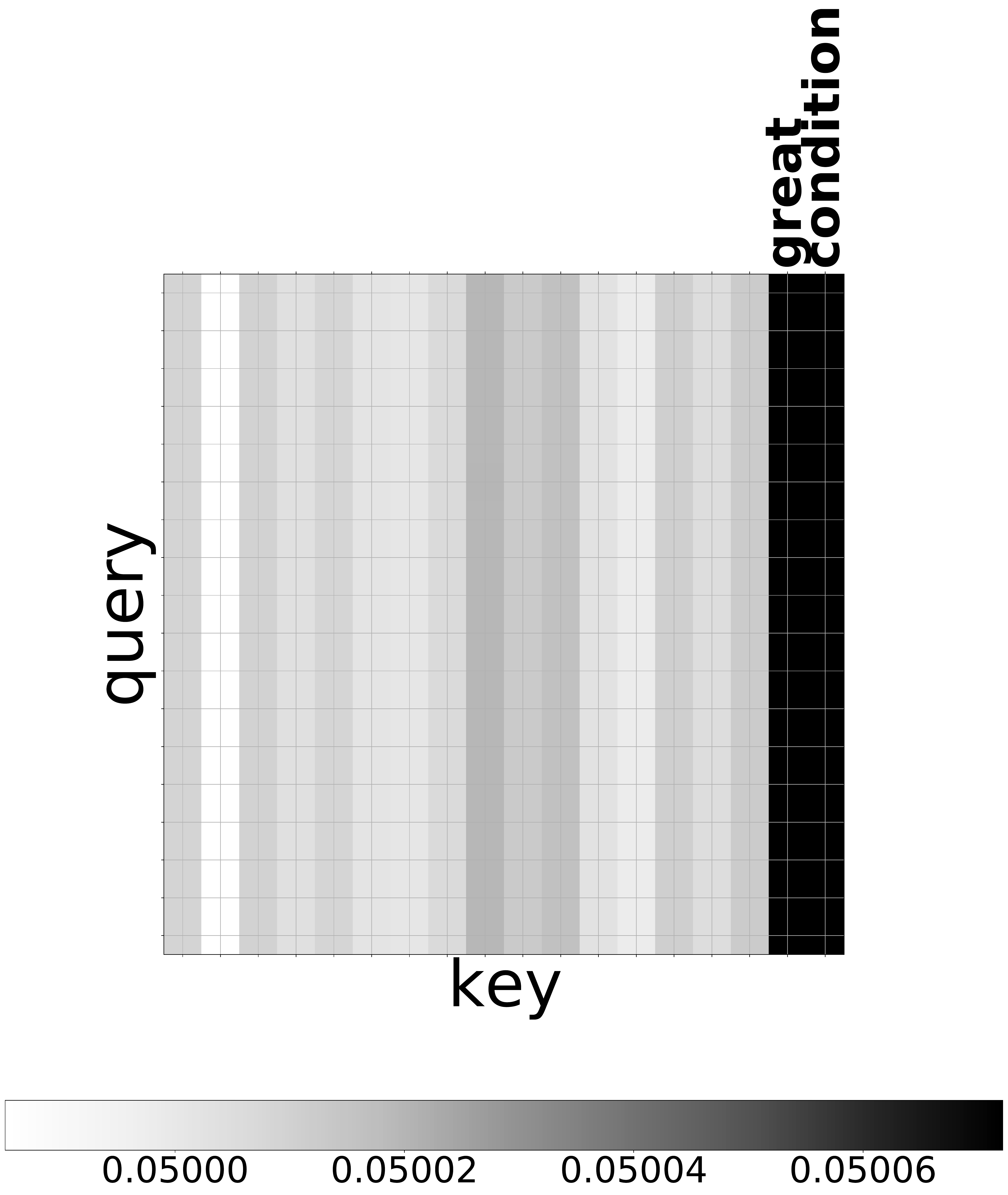}}
\caption{ClinicalBERT attention mechanism visualization. The x-axis are the query tokens and the y-axis are the key tokens. Panels (a) and (b) are two head attention mechanisms for a patient. The input notes read ``remain intubated and feel periodically very painful with back pain while awake during mechanical ventilation”. Panels (a) and (b) extract ``back pain” and ``very painful” as prominent patterns from the two heads, respectively, which provides insight on the patient's critically ill condition. Similarly, panels (c) and (d) are two head attention mechanisms for a patient without sepsis. The input notes include ``feel comfortable and tolerating cpap well and vital signs keep stable overall great condition”. ``Well” and ``stable” stand out in panel (c) and ``great condition” emerges in panel (d). All of those words  are strong indications that the patient is in a relatively benign condition.} \label{vis}
\end{figure}

\subsection{Case Studies}
We perform case studies to evaluate the uniqueness of each modality in which they may contain information that is inaccessible by the other modality. Figure~\ref{vis} depicts four self-attention mechanisms in our model which help to understand patterns in the clinical notes. In all of the panels, the x-axis represents the query tokens and the y-axis represents the key tokens. In panels (a) and (b), we analyze the medical note ``remain
intubated and feel periodically very painful with back pain while awake during mechanical ventilation” from a patient with sepsis. Panels (a) and (b) are two different head attention mechanisms. Panel (a) indicates ``back pain” and panel (b) extracts ``very painful” as prominent patterns, respectively. Similarly, panels (c) and (d) are two head attention mechanisms for a patient that ends up with no sepsis. The input note is ``feel comfortable and tolerating cpap well and vital signs keep stable overall great condition”.  CNM finds ``well”,  ``stable” and ``great condition” in panels (c) and (d), respectively. Both ``very painful” and ``great condition” help in understanding the patients' conditions and strongly correlate with the final sepsis outcomes. The indications from extracted patterns to patient outcomes show the effectiveness of the ClinicalBERT representations for clinical notes.

Then, we compare some physiological feature values from MPTS data in Figure~\ref{fig_case}, which plots the univariate distributions of selected features for sepsis and non-sepsis patients, respectively.  The orange and blue curves are density curves of observed components of features. The orange curves represent the density curves of sepsis, and the blue ones represent no sepsis. Dashed vertical lines are two patients' corresponding 
measurement values, who were correctly classified by the proposed model (PTSM + CNM) while misclassified by CNM only. Case studies suggest that single modality does not contain all the possible information that benefits the final prediction. Consequently, using both MPTS data and clinical notes can help obtain more information, which is conducive to the better predictive performance of the model.

\section{Conclusion}\label{conclusion}

In this paper, we incorporate multivariate physiological time series data and clinical notes with Transformer for early prediction of sepsis. Comprehensive experiments are conducted on two large critical care datasets, including baseline comparison, ablation analysis, and case studies. Our results demonstrate the effectiveness of our method when using both data modalities, which consistently outperforms competitive baselines on all metrics. Further analysis, and specifically to include clinicians' treatment measures in the input data, are worth exploring.

\section{Acknowledgments}
This work was funded by the National Institutes for Health (NIH) grant
NIH 7R01HL149670.

%
%
%

\begin{thebibliography}{}

\bibitem{singer2016third}
Singer, M., Deutschman, C.S., Seymour, C.W., et al.: The third international consensus definitions for sepsis and septic shock (Sepsis-3). Jama, vol. 315, pp. 801-810 (2016).

\bibitem{liu2014hospital}
Liu, V., Escobar, G.J., Greene, J.D., et al.: Hospital deaths in patients with sepsis from 2 independent cohorts. Jama, vol. 312, pp. 90-92 (2014).

\bibitem{desautels2016prediction}
Desautels, T., Calvert, J., Hoffman, J., et al.: Prediction of sepsis in the intensive care unit with minimal electronic health record data: a machine learning approach. JMIR medical informatics, vol. 4, pp. e5909 (2016).

\bibitem{saqib2018early}
Saqib, M., Sha, Y., Wang, M. D.: Early prediction of sepsis in EMR records using traditional ML techniques and deep learning LSTM networks. In: 40th Annual International Conference of the IEEE Engineering in Medicine and Biology Society (EMBC), pp. 4038-4041 (2018).

\bibitem{masino2019machine}
Masino, A. J., Harris, M. C., Forsyth, D., et al.: Machine learning models for early sepsis recognition in the neonatal intensive care unit using readily available electronic health record data. PloS one, vol. 14, pp. e0212665 (2019).

\bibitem{lipton2016directly}
Lipton, Z. C., Kale, D., Wetzel, R.: Directly modeling missing data in sequences with rnns: Improved classification of clinical time series. In: Machine learning for healthcare conference, pp. 253-270 (2016).

\bibitem{feng2020explainable}
Feng, J., Shaib, C., Rudzicz, F.: Explainable clinical decision support from text. In: Proceedings of the 2020 Conference on Empirical Methods in Natural Language Processing (EMNLP), pp. 1478-1489 (2020).

\bibitem{johnson2016mimic}
Johnson, A. E., Pollard, T. J., Shen, L., et al.: MIMIC-III, a freely accessible critical care database. Scientific data, vol. 3, pp. 1-9 (2016).

\bibitem{pollard2018eicu}
Pollard, T. J., Johnson, A. E., Raffa, J. D., et al.: The eICU Collaborative Research Database, a freely available multi-center database for critical care research. Scientific data, vol. 5, pp. 1-13 (2018).

\bibitem{patrick2010high}
Patrick, J., Li, M.: High accuracy information extraction of medication information from clinical notes: 2009 i2b2 medication extraction challenge. Journal of the American Medical Informatics Association, vol. 17, pp. 524-527 (2010).

\bibitem{deleger2013large}
Deleger, L., Molnar, K., Savova, G., et al.: Large-scale evaluation of automated clinical note de-identification and its impact on information extraction. Journal of the American Medical Informatics Association, vol. 20, pp. 84-94 (2013).

\bibitem{goodwin2016medical}
Goodwin, T. R., Harabagiu, S. M.: Medical question answering for clinical decision support. In: Proceedings of the 25th ACM international on conference on information and knowledge management, pp. 297-306 (2016).

\bibitem{devlin2018bert}
Devlin, J., Chang, M. W., Lee, K., et al.: Bert: Pre-training of deep bidirectional transformers for language understanding. arXiv preprint arXiv:1810.04805 (2018).

\bibitem{mikolov2013distributed}
Mikolov, T., Sutskever, I., Chen, K., et al.: Distributed representations of words and phrases and their compositionality. In: Advances in neural information processing systems, pp. 3111-3119 (2013).

\bibitem{pennington2014glove}
Pennington, J., Socher, R., Manning, C. D.: Glove: Global vectors for word representation. In: Proceedings of the 2014 conference on empirical methods in natural language processing (EMNLP), pp. 1532-1543 (2014).

\bibitem{lee2020biobert}
Lee, J., Yoon, W., Kim, S., et al.: BioBERT: a pre-trained biomedical language representation model for biomedical text mining. Bioinformatics, vol. 36, pp. 1234-1240 (2020).

\bibitem{alsentzer2019publicly}
Alsentzer, E., Murphy, J. R., Boag, W., et al.: Publicly available clinical BERT embeddings. arXiv preprint arXiv:1904.03323 (2019).

\bibitem{liu2013modeling}
Liu, Z., Wu, L., Hauskrecht, M.: Modeling clinical time series using gaussian process sequences. In: Proceedings of the 2013 SIAM International Conference on Data Mining, pp. 623-631 (2013).

\bibitem{liu2015clinical}
Liu, Z., Hauskrecht, M.: Clinical time series prediction: Toward a hierarchical dynamical system framework. Artificial intelligence in medicine, vol. 65, pp. 5-18 (2015).

\bibitem{lipton2015learning}
ipton, Z. C., Kale, D. C., Elkan, C., et al.: Learning to diagnose with LSTM recurrent neural networks. arXiv preprint arXiv:1511.03677 (2015).

\bibitem{song2018attend}
Song, H., Rajan, D., Thiagarajan, J. J., et al.: Attend and diagnose: Clinical time series analysis using attention models. In: Thirty-second AAAI conference on artificial intelligence (2018).

\bibitem{tsai2018learning}
Tsai, Y. H. H., Liang, P. P., et al.: Learning factorized multimodal representations. arXiv preprint arXiv:1806.06176 (2018).

\bibitem{baltruvsaitis2018multimodal}
Baltrušaitis, T., Ahuja, C., Morency, L. P.: Multimodal machine learning: A survey and taxonomy. IEEE transactions on pattern analysis and machine intelligence, vol. 41, pp. 423-443 (2018).

\bibitem{xu2018raim}
Xu, Y., Biswal, S., Deshpande, S. R., et al.: Raim: Recurrent attentive and intensive model of multimodal patient monitoring data. In: Proceedings of the 24th ACM SIGKDD international conference on Knowledge Discovery \& Data Mining, pp. 2565-2573 (2018).

\bibitem{rajkomar2018scalable}
Rajkomar, A., Oren, E., Chen, K., et al.: Scalable and accurate deep learning with electronic health records. NPJ Digital Medicine, vol. 1, pp. 1-10 (2018).

\bibitem{zhao2021bertsurv}
Zhao, Y., Hong, Q., Zhang, X., et al.: BERTSurv: BERT-Based Survival Models for Predicting Outcomes of Trauma Patients. arXiv preprint arXiv:2103.10928 (2021).

\bibitem{vaswani2017attention}
Vaswani, A., Shazeer, N., Parmar, N., et al.: Attention is all you need. In: Advances in neural information processing systems, pp. 5998-6008 (2017).

\bibitem{huang2019clinicalbert}
Huang, K., Altosaar, J., \& Ranganath, R.: Clinicalbert: Modeling clinical notes and predicting hospital readmission. arXiv preprint arXiv:1904.05342 (2019).

\bibitem{kim-2014-convolutional}
Kim, Y.: Convolutional Neural Networks for Sentence Classification. In: Proceedings of the 2014 Conference on Empirical Methods in Natural Language Processing (EMNLP), pp. 1746-1751 (2014).

\bibitem{trask2015modeling}
Trask, A., Gilmore, D., Russell, M.: Modeling order in neural word embeddings at scale. In: International Conference on Machine Learning, pp. 2266-2275 (2015).

\bibitem{harutyunyan2019multitask}
Harutyunyan, H., Khachatrian, H., Kale, D. C., et al.: Multitask learning and benchmarking with clinical time series data. Scientific data, vol. 6, pp. 1-18 (2019).

\bibitem{angus2001epidemiology}
Angus, D. C., Linde-Zwirble, W. T., Lidicker, J., et al.: Epidemiology of severe sepsis in the United States: analysis of incidence, outcome, and associated costs of care. Critical care medicine, vol. 29, pp. 1303-1310 (2001).

\bibitem{paszke2019pytorch}
Paszke, A., Gross, S., Massa, F., et al.: Pytorch: An imperative style, high-performance deep learning library. In: Advances in neural information processing systems, pp. 8026-8037 (2019).

\bibitem{kingma2014adam}
Kingma, D. P., Ba, J.: Adam: A method for stochastic optimization. arXiv preprint arXiv:1412.6980 (2014).

\bibitem{hochreiter1997long}
Hochreiter, S., Schmidhuber, J.: Long short-term memory. Neural computation, vol. 9, pp. 1735-1780 (1997).

\bibitem{schuster1997bidirectional}
Schuster, M., Paliwal, K. K.: Bidirectional recurrent neural networks. IEEE transactions on Signal Processing, vol. 45, pp. 2673-2681 (1997).

\bibitem{cho2014properties}
Cho, K., Van Merriënboer, B., Bahdanau, D., et al.: On the properties of neural machine translation: Encoder-decoder approaches. arXiv preprint arXiv:1409.1259 (2014).

\bibitem{bojanowski2017enriching}
Bojanowski, P., Grave, E., Joulin, A., et al.: Enriching word vectors with subword information. Transactions of the Association for Computational Linguistics, vol. 5, pp. 135-146 (2017).

\bibitem{peters2018deep}
Peters, M. E., Neumann, M., Iyyer, M., et al.: Deep contextualized word representations. In: Proceedings of the 2018 Conference of the North American Chapter of the Association for Computational Linguistics: Human Language Technologies, vol. 1, pp. 2227-2237 (2018).

\end{thebibliography}
%

{}

\end{document}